\pdfoutput=1
\documentclass[11pt]{article}
\usepackage{acl}

\usepackage{times}
\usepackage{latexsym}
\usepackage[T1]{fontenc}
\usepackage[utf8]{inputenc}
\usepackage{microtype}
\usepackage{inconsolata}
\usepackage{array}
\usepackage{longtable}
\usepackage{graphicx}
\usepackage{adjustbox}
\usepackage{array}
\usepackage{tcolorbox}

\usepackage{tikz}
\usetikzlibrary{trees}
\usetikzlibrary{arrows,shapes,positioning,shadows,trees}

\newcolumntype{C}[1]{>{\centering\arraybackslash}m{#1}}

\makeatletter
\def\@fnsymbol#1{}
\makeatother

\title{

Artificial Intelligence for Public Health Surveillance in Africa: Applications and Opportunities

}

\renewcommand{\footnotesize}{\fontsize{6.5}{6.5}\selectfont}

\author{\fontsize{9.45}{9.45}\selectfont Jean Marie Tshimula$^{1,2,3}$ Mitterrand Kalengayi$^{4}$ Dieumerci Makenga$^{3}$ Dorcas Lilonge$^{3}$ Marius Asumani$^{3}$ \\ \fontsize{9.45}{9.45}\selectfont \bf Déborah Madiya$^{3}$ Élie Nkuba Kalonji$^{4}$ Hugues Kanda$^{1}$ René Manassé Galekwa$^{1,3,5}$ Josias Kumbu$^{6}$ \\ \fontsize{9.45}{9.45}\selectfont \bf Hardy Mikese$^{7}$ Grace Tshimula$^{3}$ Jean Tshibangu Muabila$^{1,8}$ Christian N. Mayemba$^{1}$ D'Jeff K. Nkashama$^{1,2}$ \\ \fontsize{9.45}{9.45}\selectfont \bf Kalonji Kalala$^{1,9}$ Steve Ataky$^{10}$ Tighana Wenge Basele$^{1,3,11}$  Mbuyi Mukendi Didier$^{1,3,12,13}$ Selain K. Kasereka$^{3}$  \\ \fontsize{9.45}{9.45}\selectfont  \bf Maximilien V. Dialufuma$^{1,14}$ Godwill Ilunga Wa Kumwita$^{15}$ Lionel Muyuku$^{16}$  Jean-Paul Kimpesa$^{17}$ \\ \bf \fontsize{9.45}{9.45}\selectfont Dominique Muteba$^{18}$ Aaron Aruna Abedi$^{18}$ Lambert Mukendi Ntobo$^{19}$ Gloria M. Bundutidi$^{20}$ \\ \bf \fontsize{9.45}{9.45}\selectfont Désiré Kulimba Mashinda$^{21}$ Emmanuel Kabengele Mpinga$^{22}$ Nathanaël M. Kasoro$^{3}$ \\
\thanks{$^1$Groupe de Recherche de Prospection et Valorisation des Données (Greprovad) $^{2}$Department of Computer Science, Université de Sherbrooke, Canada $^{3}$Department of Mathematics, Statistics and Computer Science, University of Kinshasa, DRC $^{4}$University of Mbuji Mayi, DRC $^{5}$University of Klagenfurt, Austria $^6$Université Pédagogique Nationale, DRC $^{7}$Institut Supérieur Pédagogique de Kikwit, DRC $^{8}$LISV-UVSQ, Université Paris-Saclay, France $^{9}$EECS, University of Ottawa, Canada $^{10}$SyntheseAI Inc., Canada $^{11}$Karlstad University, Sweden $^{12}$Biomedical Research Unit, Hospital Monkole, Kinshasa, DRC $^{13}$University of Florida, USA $^{14}$Montreal Behavioural Medicine Centre, Centre Intégré Universitaire de Santé et Services Sociaux du Nord-de-l’Île-de-Montréal (CIUSSS-NIM), Canada $^{15}$Université Nouveaux Horizons, DRC $^{16}$University of Lorraine, France $^{17}$Centre Mère et Enfants de Barumbu and Ministry of Public Health of the DRC $^{18}$Centre National d'Intelligence Épidémiologique (CNIEP), Direction of Epidemiological Surveillance, Ministry of Public Health of the DRC $^{19}$Programme National de Lutte Contre la Trypanosomiase Humaine Africaine (PNLTHA), Direction of Epidemiological Surveillance, Ministry of Public Health of the DRC $^{20}$Graduate School of Biomedical Sciences, Nagasaki University, Japan $^{21}$School of Public Health, University of Kinshasa, DRC $^{22}$Institute of Global Health, Faculty of Medicine, University of Geneva, Switzerland. Correspondence email: \href{mailto:jeanmarie.tshimula@unikin.ac.cd}{\texttt{jeanmarie.tshimula@unikin.ac.cd}} and \href{mailto:christian.mayemba@greprovad.org}{\texttt{christian.mayemba@greprovad.org}}}
} 

\begin{document}
\maketitle
\begin{abstract}


Artificial Intelligence (AI) is revolutionizing various fields, including public health surveillance. In Africa, where health systems frequently encounter challenges such as limited resources, inadequate infrastructure, failed health information systems and a shortage of skilled health professionals, AI offers a transformative opportunity. This paper investigates the applications of AI in public health surveillance across the continent, presenting successful case studies and examining the benefits, opportunities, and challenges of implementing AI technologies in African healthcare settings. Our paper highlights AI’s potential to enhance disease monitoring and health outcomes, and support effective public health interventions. The findings presented in the paper demonstrate that AI can significantly improve the accuracy and timeliness of disease detection and prediction, optimize resource allocation, and facilitate targeted public health strategies. Additionally, our paper identified key barriers to the widespread adoption of AI in African public health systems and proposed actionable recommendations to overcome these challenges. 

\end{abstract}

\section{Introduction}

The rapid advancement of artificial intelligence (AI) technologies has created significant opportunities for improving public health surveillance and disease control across the globe. In Africa, where public health challenges are often compounded by resource limitations, and failed health information systems, AI presents a transformative potential for improving disease detection, prediction, and response. The continent faces a high burden of infectious diseases such as  tuberculosis (TB), malaria, HIV/AIDS ({\it Human immunodeficiency virus / Acquired immunodeficiency syndrome}), and Ebola, alongside emerging health threats including the Zika virus and COVID-19. Traditional methods of disease surveillance and response, while valuable, often struggle to keep pace with the dynamic and complex epidemiological landscape \cite{mbunge2023application, worsley2022strengthening}.

AI's ability to analyze vast datasets from diverse sources such as electronic health records, social media, environmental sensors, and genomic data enables the identification of patterns and anomalies indicative of disease outbreaks. Machine learning techniques, including random forests, support vector machines, and deep learning models, have shown remarkable success in providing quicker and more accurate predictions compared to conventional methods \cite{wahl2018artificial}. These technologies not only facilitate targeted interventions but also improve the efficiency of healthcare delivery in resource-limited settings \cite{lamichhane2022improved}.

In sub-Saharan Africa, AI has shown considerable promise in improving TB detection and management. Various machine learning models were developed to predict TB incidences and drug resistance, and boost diagnosis accuracy through computer-aided detection systems \cite{siamba2023application, qin2024computer, oloko2022systematic}. These technological advancements are pivotal in tackling the region's high TB burden, as they enable early detection and ensure effective treatment, which are critical for controlling the spread of the disease \cite{kuang2022accurate}.

In the context of HIV detection and prediction, AI has demonstrated significant promise in sub-Saharan Africa. Studies leveraging algorithms like XGBoost and artificial neural networks have achieved high accuracy in identifying HIV-positive individuals and predicting drug resistance mutations \cite{mutai2021use, ebulue2024developing}. This capability is crucial for developing personalized treatment plans and optimizing the allocation of healthcare resources. Similarly, predictive models for diseases such as cholera and Ebola have integrated socioeconomic, environmental, and climatic data to achieve high accuracy and inform public health strategies and resource allocation \cite{zheng2022cholera,charnley2022drought,kaseya2024climate,siettos2015modeling,zhang2015large,colubri2019machine,Hwang2023machine}.

\begin{figure*}
    \centering
    \includegraphics[width=.725\linewidth]{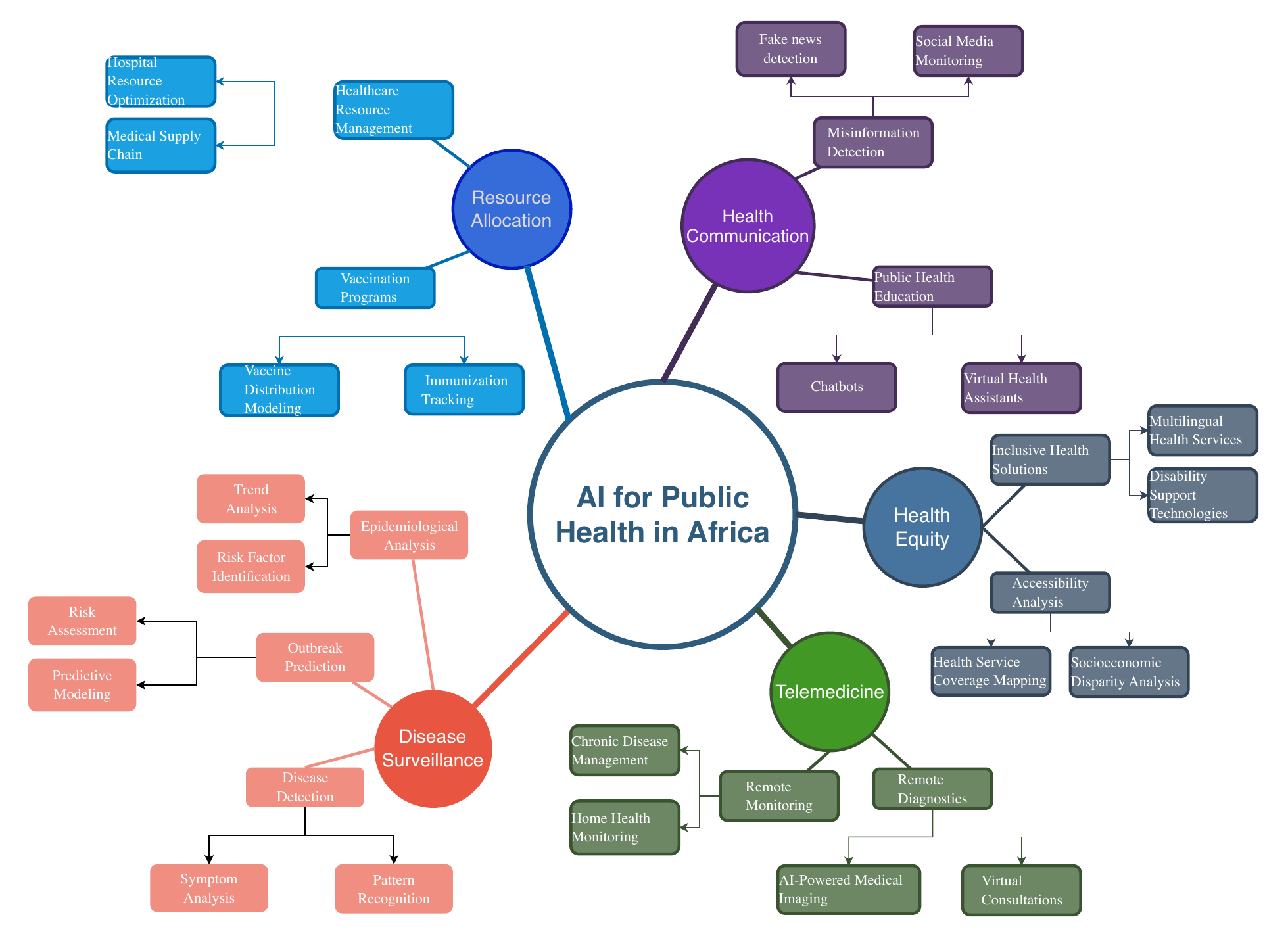}
    \caption{Taxonomy of AI applications for public health.}
    \label{fig:taxonomy}
\end{figure*}

Figure~\ref{fig:taxonomy} displays AI applications for public health and demonstrates the diverse and multifaceted roles AI can play in enhancing public health infrastructure, improving data management, supporting epidemiological research, optimizing resource allocation, fostering effective health communication, aiding policy development, facilitating remote healthcare delivery, and promoting health equity across African regions. The application of AI extends beyond infectious disease surveillance to real-time monitoring of various health conditions. For instance, AI-powered tools have been used in influenza surveillance, integrating Google search data with historical illness data to improve prediction accuracy \cite{nsoesie2021forecasting, olukanmi2021utilizing, steffen2012improving}. In the realm of vector-borne diseases, machine learning models have been developed to predict malaria prevalence based on climatic factors and achieve significant accuracy across different African regions \cite{ileperuma2023predicting, mariki2022combining, hemachandran2023performance, nkiruka2021prediction, masinde2020africa}. These models facilitate timely interventions by forecasting outbreaks and identifying high-risk zones.

Moreover, AI-driven tools have revolutionized poliovirus surveillance in Africa and significantly improved the detection and reporting of acute flaccid paralysis cases. Geographic information systems combined with AI algorithms have improved real-time mapping and analysis of polio cases, vaccination coverage, and population movements, optimizing resource allocation for vaccination campaigns \cite{shuaib2018avadar, kamadjeu2009tracking, dougherty2019paper, hamisu2022characterizing}. These systems contribute to the improvement of polio surveillance and eradication efforts.

Mental health monitoring is another area where AI has shown potential, particularly in identifying predictors of mental health disorders among healthcare workers and the general population. Machine learning algorithms have been employed to predict depression, anxiety, and stress, and identify other signals relevant to mental stated for implementing targeted interventions and providing mental health care \cite{njoroge2023use, alharahsheh2021predicting, ugar2024designing, kemp2020patient, mokhele2019prevalence, commander2020predictors, bantjes2016symptoms, bigna2019epidemiology, amu2021prevalence, ojagbemi2022pre, saal2018utility, ademola2019prevalence, simbayi2007internalized, khumalo2022measuring, muhammed2017predictors, kaputu2021ptsd}.


Tropical disease diagnosis has seen significant advancements in machine learning applications. Diseases like dengue fever, malaria, and tuberculosis, particularly vector-borne illnesses, have been the primary focus of many studies in this field. However, a common challenge persists: the lack of comprehensive data, which impacts the models' generalizability and effectiveness \cite{attai2022systematic}. To improve the reliability of machine learning in tropical disease diagnosis, there is a pressing need for more extensive and diverse datasets.


Unlike AI-based methods, traditional surveillance systems often rely on health professionals to report infections, a process that can be slow and labor-intensive, especially in low-income countries with weak health systems. AI can overcome these limitations by providing real-time data processing, reducing the time lag between the onset of an epidemic and its detection. This early detection capability is crucial for implementing timely interventions, such as case isolation and contact tracing, which can significantly curb the spread of diseases. AI-based epidemic intelligence systems such as ProMED-mail, HealthMap, Epidemic Intelligence from Open Sources (EIOS), BlueDot, Metabiota, the Global Biosurveillance Portal, Epitweetr, and EPIWATCH, have demonstrated the ability to filter and curate data from diverse sources, which derive signals relevant to early epidemic detection and response \cite{macintyre2023artificial}.

This paper investigates the diverse applications of AI in public health surveillance across Africa, highlights successful case studies, and identifies opportunities for further innovation. Specifically, this paper aims to answer two key research questions: {\bf RQ1:} {\it How can artificial intelligence improve disease detection, prediction, and surveillance in public health in low-resource regions?} {\bf RQ2:} {\it What are the key challenges and barriers to implementing AI technologies in public health systems in low-resource regions, and what strategies can overcome these challenges?}.

The integration of AI-driven predictive models into public health systems can significantly improve disease outbreak preparedness and response, ultimately save lives and reduce the disease burden. While there are numerous public health concerns in Africa, this paper focuses exclusively on the following: {\it HIV, cholera, Ebola, measles, tuberculosis, influenza, Zika virus, COVID-19, malaria, poliovirus}, and {\it mental health}.

The remainder of this work is organized as follows. Section \S\ref{app_ai} discusses the applications of artificial intelligence in public health surveillance and highlights various case studies and methodologies. Section \S\ref{opp_ai} focuses on the opportunities presented by AI for improving healthcare delivery and disease monitoring.  Section \S\ref{challenge} examines the challenges and barriers to implementing AI technologies in low-resource settings, and addresses the ethical considerations surrounding the use of AI in public health. Section \S\ref{discussion} provides a discussion on the implications of AI integration in public health systems. Section \S\ref{fdfuturedirection} explores future directions for research and development in this field. Finally, Section \S\ref{conclusion} concludes the paper with reflections on the potential of AI to revolutionize public health surveillance and improve health outcomes in low-resource regions.

\section{Applications of AI in public health surveillance} \label{app_ai}

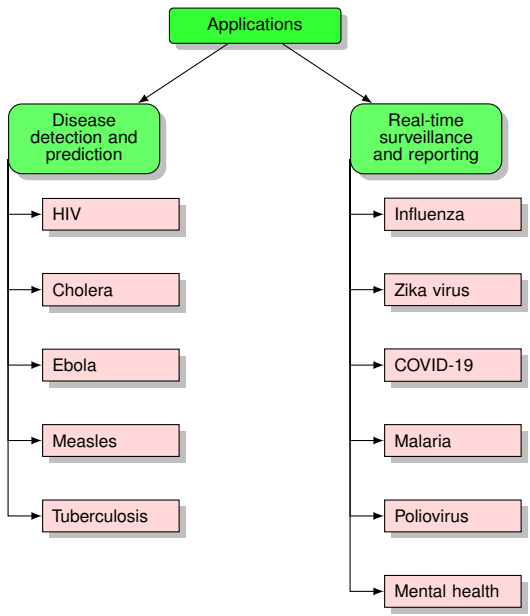
\begin{figure}[h!]
    \centering
    \tikzset{
    basic/.style  = {draw, text width=2cm, drop shadow,
    font=\sffamily\footnotesize, rectangle},
    root/.style   = {basic, rounded corners=1.5pt, thin, align=center,
                   fill=green!80, font=\sffamily\footnotesize},
                   level 2/.style = {basic, rounded corners=5pt, thin, align=center, fill=green!60,
                   text width=4.5em, font=\sffamily\footnotesize},
                   level 3/.style = {basic, thin, align=left, fill=pink!60, text width=4em, font=\sffamily\footnotesize}
    }
    \begin{tikzpicture}[
  level 1/.style={sibling distance=45mm},
  edge from parent/.style={->,draw},
  >=latex]

\node[root] {Applications}
  child {node[level 2] (c2) {Disease detection and prediction}}
  child {node[level 2] (c3) {Real-time surveillance and reporting}};

\begin{scope}[every node/.style={level 3}]
\node [below of = c2, xshift=10pt] (c21) {HIV};
\node [below of = c21] (c22) {Cholera};
\node [below of = c22] (c23) {Ebola};
\node [below of = c23] (c24) {Measles};
\node [below of = c24] (c25) {Tuberculosis};

\node [below of = c3, xshift=10pt] (c31) {Influenza};
\node [below of = c31] (c32) {Zika virus};
\node [below of = c32] (c33) {COVID-19};
\node [below of = c33] (c34) {Malaria};
\node [below of = c34] (c35) {Poliovirus};
\node [below of = c35] (c36) {Mental health};
\end{scope}

\foreach \value in {1,...,5}
  \draw[->] (c2.195) |- (c2\value.west);

\foreach \value in {1,...,6}
  \draw[->] (c3.195) |- (c3\value.west);
\end{tikzpicture}
\caption{Applications of AI in public health surveillance discussed in this paper.}\label{taxonomy_app}
    
\end{figure}

\begin{figure}[t!]
    \centering
    \includegraphics[width=0.45\textwidth]{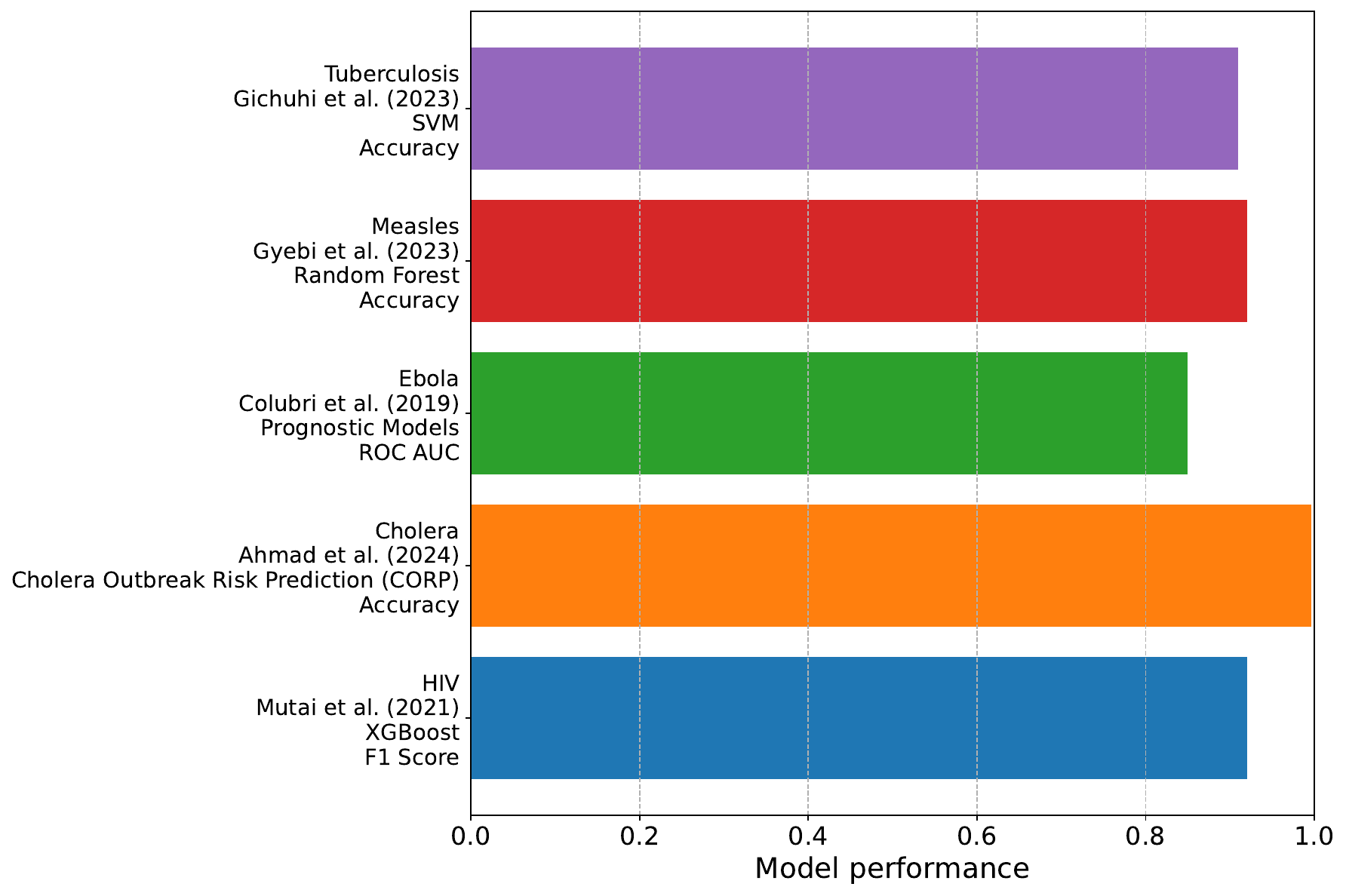} 
    \caption{Performance metrics of best models per disease for Disease prediction and detection. Note that {\it y-axis} indicates {\it Disease + Authors + Model + Metric}.}
    \label{fig:model_performance_histo_disease}
\end{figure}

Artificial Intelligence (AI) revolutionizes public health surveillance by improving disease detection and prediction capabilities, as well as real-time surveillance and reporting. In this paper, we categorize the applications of AI in public health into two main areas (Figue \ref{taxonomy_app}): disease detection and prediction (\S\ref{detection_prediction}, Table \ref{tab:disease_detection_prediction}, Figure \ref{fig:model_performance_histo_disease}), and real-time surveillance and reporting (\S\ref{realtime}, Table \ref{tab:real_time_surveillance}, Figure \ref{model_performance_histo_real_time_surv}). In the following sections, we investigate these applications in detail and examine advancements and the impact of AI on public health strategies in Africa and beyond ({\bf RQ1}).

\subsection{Disease detection and prediction} \label{detection_prediction}
AI's potential to revolutionize disease detection and prediction is evident through its application in analyzing vast datasets from sources like electronic health records, social media, and environmental sensors. Machine learning algorithms can help identify patterns and anomalies indicative of disease outbreaks, and provide quicker and more accurate predictions than traditional methods. For instance, in HIV detection and prediction, machine learning has demonstrated significant promise in sub-Saharan Africa. Machine learning models helped identify HIV-positive individuals and predict drug resistance mutations with high accuracy \cite{mutai2021use,ebulue2024developing} (\S\ref{hiv}). Additionally, machine learning models have been developed to predict HIV outbreaks by identifying high-risk populations, and improving retention and viral load suppression in HIV treatment programs demographic data, and biological data from screenings \cite{xu2022machine,powers2018prediction,maskew2022applying}. This predictive capability is crucial for public health authorities to efficiently allocate resources and implement targeted interventions \cite{chikusi2022machine,tim2006predicting,oladokun2019predicting,ebulue2024machine,alie2024machine,roche2024measuring}.

Similarly, machine learning techniques have been employed to predict and monitor cholera and Ebola outbreaks (\S\ref{cholera}, \S\ref{ebola}). In sub-Saharan Africa, predictive models for cholera outbreaks have achieved remarkable accuracy by integrating socioeconomic, environmental, and climatic data \cite{zheng2022cholera,charnley2022drought,kaseya2024climate}. These models have informed public health strategies and resource allocation, particularly in regions vulnerable to cholera due to inadequate sanitation and water access \cite{ahmad2024can,Buechner2022drought,leo2019machine}. For Ebola, machine learning models have utilized patient data, geodemographics, and environmental factors to predict outbreaks and assess intervention effectiveness \cite{siettos2015modeling,zhang2015large,colubri2019machine}. These approaches have facilitated early detection, timely interventions, and improved outbreak management \cite{Hwang2023machine,pigott2014mapping,viboud2018rapidd,buscema2020analysis}. Machine learning's ability to analyze complex data and generate accurate predictions is also being explored for other diseases like measles (\S\ref{measles}), where predictive models have demonstrated potential in enhancing surveillance and immunization strategies across various African countries \cite{gyebi2023prediction,michael2024trends,nsubuga2017positive,leung2024combining,thakkar2024estimating}. Integrating AI-driven predictive models into public health systems can significantly improve disease outbreak preparedness and response, and save lives and reduce disease burden.

For tuberculosis, machine learning models have improved detection and diagnosis accuracy, and predicted incidences and drug resistance in sub-Saharan Africa \cite{siamba2023application, qin2024computer, oloko2022systematic} (\S\ref{tuberculosis}). These advancements are vital for early detection and effective treatment of tuberculosis, helping to control its spread \cite{kuang2022accurate}.

\begin{table*}[h!]
\centering
\fontsize{5.15}{5.15}\selectfont
\renewcommand{\arraystretch}{1.2}
\caption{Summary of AI Applications in Disease Detection and Prediction}
\begin{tabular}{|C{2cm}|C{2.5cm}|C{1cm}|C{2cm}|C{2.cm}|C{3.5cm}|}
\hline
\textbf{Authors} & \textbf{Model} & \textbf{Disease} & \textbf{Country} & \textbf{Application} & \textbf{Case Study Description} \\ \hline
\citet{mutai2021use} & XGBoost & HIV & Sub-Saharan Africa & HIV detection & Used socio-behavioral data to improve HIV-positive identification with high accuracy. \\ \hline
\citet{ebulue2024developing} & Deep learning, ensemble & HIV & Sub-Saharan Africa & HIV drug resistance prediction & Leveraged genomic data to predict HIV drug resistance mutations, enhancing personalized medicine. \\ \hline
\citet{xu2022machine} & Machine learning & HIV & Sub-Saharan Africa & HIV risk prediction & Discussed the benefits and limitations of machine learning for accurate HIV risk prediction. \\ \hline
\citet{powers2018prediction} & Logistic regression & HIV & Sub-Saharan Africa & HIV high viremia prediction & Developed risk score algorithms to identify individuals at risk of extended high viremia. \\ \hline
\citet{maskew2022applying} & Logistic regression, RF, AdaBoost & HIV & South Africa & HIV treatment program outcomes & Predicted retention and viral load suppression in HIV treatment programs using patient-level data. \\ \hline
\citet{chikusi2022machine} & Random Forest, XGBoost, ANN & HIV & Tanzania & HIV index testing & Improved prediction and visualization of HIV index testing using machine learning models. \\ \hline
\citet{tim2006predicting} & Artificial Neural Network & HIV & South Africa & HIV status classification & Used demographic data to classify HIV status with high accuracy. \\ \hline
\citet{oladokun2019predicting} & Decision Tree, Logistic Regression & HIV & South Africa & HIV status prediction & Compared machine learning models for predicting HIV status based on demographic data. \\ \hline
\citet{ebulue2024machine} & Machine learning & HIV & Sub-Saharan Africa & HIV outbreak prediction & Forecasted HIV outbreaks by identifying high-risk populations using demographic and epidemiological data. \\ \hline
\citet{alie2024machine} & J48 Decision Tree & HIV & Ethiopia & Adolescent HIV testing prediction & Analyzed adolescent data to predict HIV testing behaviors with high accuracy. \\ \hline
\citet{roche2024measuring} & AI algorithm & HIV & Kenya & HIV self-test result interpretation & Improved quality assurance of HIV self-tests by comparing AI interpretations with human interpretations. \\ \hline
\citet{zheng2022cholera} & Machine learning & Cholera & Sub-Saharan Africa & Cholera outbreak prediction & Analyzed outbreak data to predict and monitor cholera outbreaks across 25 countries. \\ \hline
\citet{charnley2022drought} & Generalized Linear Models & Cholera & Africa & Cholera outbreak risk assessment & Linked drought and environmental factors to cholera outbreak risks. \\ \hline
\citet{kaseya2024climate} & Machine learning & Cholera & Southern Africa & Cholera prevention and control & Developed strategies incorporating climatic effects to manage cholera outbreaks. \\ \hline
\citet{ahmad2024can} & Machine learning & Cholera & Nigeria & Cholera outbreak risk prediction & Developed a highly accurate Cholera Outbreak Risk Prediction model. \\ \hline
\citet{leo2019machine} & XGBoost & Cholera & Tanzania & Cholera outbreak prediction & Linked seasonal weather changes to cholera outbreaks, achieving high prediction accuracy. \\ \hline
\citet{siettos2015modeling} & Agent-based simulations & Ebola & West Africa & Ebola epidemic dynamics & Modeled the 2014 Ebola virus epidemic, emphasizing intervention impacts. \\ \hline
\citet{zhang2015large} & Machine learning & Ebola & West Africa & Ebola outbreak prediction & Predicted Ebola outbreaks using large-scale simulations based on geodemographics. \\ \hline
\citet{colubri2019machine} & Prognostic models & Ebola & Liberia, Sierra Leone & Ebola death likelihood prediction & Developed models to predict death likelihood during the 2014-16 outbreak. \\ \hline
\citet{pigott2014mapping} & Species distribution models & Ebola & Central and West Africa & Ebola zoonotic niche mapping & Mapped the zoonotic niche of EVD across 22 countries, identifying at-risk regions. \\ \hline
\citet{gyebi2023prediction} & Random Forest & Measles & Ghana & Measles case prediction & Compared machine learning techniques for predicting measles cases, with Random Forest showing superior results. \\ \hline
\citet{nsubuga2017positive} & Case-based surveillance & Measles & Uganda & Measles surveillance evaluation & Assessed the effectiveness of measles surveillance, highlighting the need for robust data collection. \\ \hline
\citet{leung2024combining} & Dynamic model & Measles & Africa & Measles reporting rate and incidence estimation & Combined clinical and diagnostic data to estimate measles incidence under varying vaccination coverages. \\ \hline
\citet{thakkar2024estimating} & Transmission model & Measles & Somalia & Measles vaccination impact assessment & Evaluated the impact of vaccination campaigns on measles incidence and mortality. \\ \hline
\citet{jahanbin2024using} & LSTM & Measles & South Africa & Measles social media analysis & Analyzed social media data to understand public sentiment and outbreak patterns. \\ \hline
\citet{maradze2021modelling} & ANN & Measles & Djibouti & Measles immunization rate forecasting & Forecasted child immunization rates using an ANN model. \\ \hline
\citet{doshi2015effect} & Integrated Disease Surveillance & Measles & DRC & Measles immunization impact assessment & Assessed the impact of immunization activities on measles incidence. \\ \hline
\citet{fatiregun2014epidemiology} & Trend analysis & Measles & Nigeria & Measles infection trends & Analyzed seasonal peaks and projected trends in measles infections. \\ \hline
\citet{ferrari2008dynamics} & Stochastic metapopulation model & Measles & Niger & Measles outbreak dynamics & Studied the episodic nature of measles outbreaks and the impact of vaccination. \\ \hline
\citet{graham2018challenges} & Statistical methods & Measles & Guinea & Measles outbreak simulation & Modeled measles outbreak following the Ebola crisis and the impact of vaccination campaigns. \\ \hline
\citet{eilertson2019estimation} & State-space model & Measles & Niger, Ethiopia, DRC & Measles transmission prediction & Developed a model to predict measles cases and underreporting rates. \\ \hline
\citet{uyar2019forecasting} & GA-based RFNN & Measles & Ethiopia & Measles case prediction & Compared GA-based RFNN and ANFIS for predicting monthly measles cases. \\ \hline
\citet{james2022mathematical} & Deterministic model & Measles & Nigeria & Measles transmission dynamics & Studied the impact of vaccination and hospitalization rates on measles transmission. \\ \hline
\citet{akinbobola2018predicting} & Poisson regression, ARIMA & Measles & Nigeria & Weather impact on measles incidence & Analyzed the relationship between weather variables and measles incidence. \\ \hline
\citet{goodson2011changing} & Statistical analysis & Measles & Africa & Measles vaccination coverage & Examined the correlation between vaccination coverage and measles case age distribution. \\ \hline
\citet{szusz2010review} & Review & Measles & Low-income countries & Measles transmission modeling & Reviewed epidemiology data essential for dynamic models of measles transmission. \\ \hline
\citet{bishai2011cost} & Dynamic stochastic model & Measles & Uganda & Cost-effectiveness of SIAs & Evaluated the cost-effectiveness of supplementary immunization activities. \\ \hline
\citet{parpia2020spatio} & Multivariate time-series models & Measles & Cameroon & Spatial dynamics of measles outbreak & Characterized spatial heterogeneity in vaccination coverage and transmission patterns. \\ \hline
\citet{lee2019economic} & Compartment model & Measles & Kenya & Measles immunization outreach strategy & Evaluated the economic benefits of a two-dose immunization strategy for hard-to-reach children. \\ \hline

\citet{siamba2023application} & ARIMA, hybrid ARIMA & Tuberculosis & Kenya & TB incidence prediction & Applied ARIMA and hybrid ARIMA models to predict TB incidences among children, demonstrating significant under-reporting. \\ \hline
\citet{ojugo2021multi} & Bayesian networks & Tuberculosis & Nigeria & TB diagnosis & Developed a decision-making framework using Bayesian networks to predict TB cases with high accuracy. \\ \hline
\citet{sekandi2023application} & 3D ResNet & Tuberculosis & Uganda & TB medication adherence & Evaluated self-recorded medication intake videos with a 3D ResNet model for real-time monitoring of TB medication adherence. \\ \hline
\citet{gichuhi2023machine} & SVM, classification algorithms & Tuberculosis & Uganda & TB treatment non-adherence & Identified individual risk factors for TB treatment non-adherence using five classification algorithms, with SVM achieving highest accuracy. \\ \hline
\citet{oshinubi2023mathematical} & Deterministic model & Tuberculosis & East Africa & TB impact analysis & Developed a mathematical model to study the impact of vaccination and treatment strategies on TB prevalence. \\ \hline

\end{tabular}
\label{tab:disease_detection_prediction}
\end{table*}

\subsubsection{HIV detection and prediction}\label{hiv}

The application of machine learning in HIV detection and prediction has garnered significant attention, particularly in sub-Saharan Africa, where the HIV epidemic remains a critical public health challenge. Various studies have explored the potential of machine learning techniques to improve HIV prevention and intervention strategies.

\citet{mutai2021use} utilized machine learning algorithms to identify HIV predictors using socio-behavioral data from the Population-based HIV Impact Assessment (PHIA) surveys conducted in four sub-Saharan African countries. Their study applied the XGBoost algorithm, which significantly improved the identification of HIV-positive individuals with an F1 score of 90\% for males and 92\% for females. The key predictive features included age, relationship with the family head, education level, and wealth quintile. This approach demonstrated the potential of machine learning to facilitate targeted screening interventions.

Similarly, \citet{ebulue2024developing} proposed a novel framework leveraging genomic data and AI to predict HIV drug resistance mutations. The research emphasizes the need for innovative strategies to manage and anticipate resistance mutations to optimize antiretroviral therapy. Integrating genomic data from HIV-infected individuals with AI algorithms, they aimed to identify genetic variations linked to drug resistance, enhancing personalized medicine in HIV care. The model training employed machine learning techniques like deep learning and ensemble methods, validated through cross-validation and independent testing. Incorporating clinical data such as treatment history and viral load improved the predictive accuracy of the models, potentially enabling clinicians to tailor treatment plans based on individual genetic profiles.

Another study by \citet{xu2022machine} highlighted the role of machine learning in HIV risk prediction. The authors discussed the benefits and limitations of using machine learning for HIV risk prediction, and they emphasized the importance of accurate and granular risk prediction for directing HIV testing campaigns and pre-exposure prophylaxis (PrEP) prescriptions. The study underscored the utility of machine learning in capturing complex interactions within data, which is crucial for effective HIV risk prediction in both high-income and resource-limited settings.

\citet{powers2018prediction} developed predictive models for identifying newly HIV-infected individuals at risk of extended high viremia (EHV) in sub-Saharan Africa. The researchers used logistic regression to create risk score algorithms based on clinical, demographic, and laboratory indicators, including sex, age, number of acute retroviral syndrome (ARS) symptoms, CD4 count, and viral load at diagnosis. The models showed excellent performance with a c-statistic of 0.80 in the overall population and similar values within specific HIV-1 subtypes (A, C, and D). The final models effectively predicted EHV with high sensitivity (85\%) and moderate specificity (61\%). These results suggest that simple risk score algorithms can reliably identify individuals likely to sustain high viral loads if treatment is delayed.

Additionally, \citet{maskew2022applying} utilized machine learning algorithms on anonymized patient-level data from HIV treatment programs in South Africa. The study aimed to predict two outcomes: retention (defined as attending the next scheduled clinic visit within 28 days) and viral load (VL) suppression. The dataset included demographic, clinical, behavioral, and laboratory data from 2016 to 2018. Three classification algorithms were evaluated: logistic regression, random forest, and AdaBoost. The study found that the model's predictive accuracy, measured by the area under the curve (AUC), ranged from 0.69 for retention to 0.76 for VL suppression. The predictors included prior late visits, the number of prior VL tests, and the duration on treatment. The results indicated that machine learning could effectively identify patients at risk of disengagement and unsuppressed VL. This capability could potentially improve targeted interventions and resource allocation in HIV treatment programs.

\citet{chikusi2022machine} focused on developing a machine learning model to improve the prediction and visualization of HIV index testing. The research gathered data from the Kilimanjaro, Arusha, and Manyara regions, comprising 6346 samples with 11 features. The dataset was split into training (5075 samples) and testing sets (1270 samples) and subjected to Random Forest (RF), XGBoost, and Artificial Neural Networks (ANN) algorithms. The RF algorithm demonstrated the lowest mean absolute error (MAE) at 1.1261, indicating superior performance compared to XGBoost (MAE of 1.2340) and ANN (MAE of 1.1268). Data visualization revealed that females (82.6\%) are more likely to test for HIV than males (17.4\%), with higher instances of partner listing in Kilimanjaro. The study underscores the potential of machine learning in predicting and visualizing HIV index testing, providing decision-makers with a tool to devise effective interventions to curb HIV spread.

\citet{tim2006predicting} investigated the use of artificial neural networks to classify HIV status based on demographic data from South African antenatal surveys. The key inputs to the neural network included age, gravidity, parity, race, and location, while the output was the binary classification of HIV status. A multilayer perceptron with a logistic activation function was trained using a cross-entropy error function to provide a probabilistic output. The neural network achieved an accuracy of 88\% in predicting HIV status, with an area under the ROC curve (AUC) of 0.85, indicating high sensitivity (85\%) and specificity (80\%). The study also explored the impact of missing data by training an auto-associative neural network to approximate missing values and then using global optimization methods to fill in these gaps. Results showed that the inclusion of imputed data improved the prediction accuracy, with AUC improving to 0.87.

In another study, \citet{oladokun2019predicting} investigated the efficacy of machine learning models in predicting HIV status based on demographic data. Using data from the 2016 DHS survey involving 7808 women aged 15-49 in South Africa, the study compared the performance of a decision tree model and logistic regression. Results indicated that the logistic regression model outperformed the decision tree model with a testing accuracy of 68.03\% compared to 63.93\% for the decision tree. The area under the curve (AUC) for the ROC curve was 0.682 for logistic regression and 0.652 for the decision tree.

\citet{ebulue2024machine} examined the effectiveness of machine learning in forecasting HIV outbreaks in regions highly impacted by the epidemic. The authors highlighted the potential of machine learning models to accurately identify patterns and trends in HIV data, leveraging demographic, epidemiological, and behavioral data to predict high-risk populations and geographical areas. This predictive capability is crucial for public health authorities to efficiently allocate resources and implement targeted interventions. The study acknowledged challenges such as data quality issues and the complexity of HIV transmission dynamics, suggesting that integrating multiple data sources and employing advanced machine learning techniques can improve model robustness.

Focusing on adolescent HIV testing, \citet{alie2024machine} utilized machine learning techniques to analyze data from 4502 adolescents in Ethiopia. They found the J48 decision tree algorithm to be the most effective, with an accuracy of 81.29\% and an ROC curve of 86.3\%. The key predictors included age, knowledge of HIV testing locations, age at first sexual encounter, recent sexual activity, and exposure to family planning.

\citet{roche2024measuring} evaluated the effectiveness of an AI algorithm in interpreting HIV self-test (HIVST) results. Conducted in Kisumu, Kenya, the study enrolled 1500 participants who used blood-based HIVST kits at private pharmacies. The AI algorithm's results were compared against interpretations by pharmacy clients, providers, and an expert panel. The AI algorithm achieved a perfect sensitivity (100\%) and negative predictive value (100\%), with a specificity of 98.8\%. In contrast, pharmacy clients and providers showed slightly lower sensitivity but perfect specificity and positive predictive value. The study concluded that AI could improve the quality assurance of HIVST, particularly in reducing false negatives, and highlighted the potential for AI to support differentiated HIV service delivery models.

\subsubsection{Cholera outbreaks}\label{cholera}

Cholera outbreaks remain a significant public health concern in Africa, particularly in regions with inadequate sanitation and unreliable access to clean water. Researchers have explored the application of machine learning techniques to predict and monitor cholera outbreaks, with notable successes. A comprehensive study analyzing cholera outbreaks in sub-Saharan Africa from 2010 to 2019 identified 999 suspected cholera outbreaks across 25 countries \cite{zheng2022cholera}. The median attack rate was 0.8 per 1000 people, with a median epidemic duration of 13 weeks. Interestingly, larger attack rates were associated with longer outbreak durations but lower case fatality risks.

Several socioeconomic and environmental factors contribute to the persistence of cholera in Africa. Drought, poor access to water, marginalization of refugees and nomadic populations, expansion of informal urban settlements, and demographic risks are significant contributors to cholera outbreaks \cite{charnley2022drought}. Climate change may exacerbate these risks by altering precipitation patterns and increasing the frequency of extreme weather events. Recent data from January 2023 to January 2024 reveal that 19 African countries reported cholera outbreaks, resulting in 252,934 cases and 4,187 deaths \cite{kaseya2024climate}. The southern Africa region has been particularly affected, with countries like Zambia and South Africa reporting high case fatality rates.

To address these challenges, African countries and regional bodies have taken steps to prioritize cholera prevention and control. The Southern Africa Development Community (SADC) has recommended that member states develop multisectoral response plans incorporating climatic effects on cholera re-emergence \cite{kaseya2024climate}. Enhancing drought diplomacy, including multi-country drought response plans and water agreements, could improve water management and resource sharing during droughts and subsequent outbreaks.

Machine learning techniques have shown great promise in cholera outbreak prediction and monitoring. For example, a study in Nigeria developed a Cholera Outbreak Risk Prediction (CORP) model using machine learning techniques, and CORP achieved an accuracy of 99.62\% \cite{ahmad2024can}. Similarly, UNICEF has leveraged AI to combat cholera in African countries, including the Democratic Republic of the Congo, Malawi, Mozambique, Zambia, and Zimbabwe \cite{Buechner2022drought}.

In Tanzania, \citet{leo2019machine} developed a machine learning model to predict cholera outbreaks by linking them to seasonal weather changes. Using adaptive synthetic sampling approach (ADASYN) for oversampling and principal component analysis (PCA) for dimensionality reduction, the XGBoost classifier emerged as the best performer and achieved a balanced accuracy score of 0.767 $\pm$ 0.09. In coastal India, \citet{campbell2020cholera} used a Random Forest classifier model trained on atmospheric, terrestrial, and oceanic satellite-derived essential climate variables, and achieved an accuracy of 0.99.

Machine learning applications in cholera outbreak prediction are not limited to Africa. Studies in Yemen and Bangladesh have also demonstrated the potential of AI for cholera outbreak prediction \cite{usmani2023combating, campbell2020cholera, emch2008local, leo2022complexity}. These approaches highlight the versatility and adaptability of machine learning models across different geographic contexts.

Further research has focused on integrating various data sources and modeling techniques to improve prediction accuracy. \citet{onyijen2023data} investigated machine learning algorithms to predict cholera outbreaks in West Africa, finding that the decision tree model exhibited the highest accuracy at 99.8\%. \citet{leo2020reference} developed a machine learning model to predict cholera outbreaks by integrating weather and health data, and achieved an overall accuracy of 78.5\% with an ensemble method.

The use of social media data has also been explored. \citet{jideani2024machine} examined emotions expressed in social media posts about the cholera outbreak in Hammanskraal, South Africa, using various machine learning models. The LSTM model outperformed others and yielded a classification accuracy of 76\%.

Several studies have highlighted the limitations of traditional models in predicting cholera outbreaks. \citet{usmani2021review} emphasized the need for comprehensive models that account for both epidemic and endemic cholera dynamics. \citet{fleming2007fuzzy} described the development of an early warning prototype tool using GIS and fuzzy logic, while \citet{nkwayep2024prediction} presented a detailed analysis of cholera transmission dynamics in Cameroon.

The integration of Internet of things (IoT) and AI technologies has also shown potential in cholera prediction. \citet{ogore2021offline} proposed an offline edge AI device for rural water taps and achieved a 94\% accuracy rate with an SVM model. Additionally, \citet{charnley2021exploring} investigated the association between drought and cholera outbreaks across Africa, using generalized linear models to predict cholera outbreaks. 

Overall, reducing population vulnerability through poverty alleviation and improving water, sanitation, and hygiene (WASH) services are crucial steps in mitigating cholera outbreaks. Strengthening health systems and increasing research on drought-related health impacts are essential for effectively addressing the persistent cholera burden in Africa.

\subsubsection{Ebola prediction}\label{ebola}

Ebola virus disease (EVD) has been a significant public health concern in Africa, particularly in sub-Saharan regions. The use of machine learning techniques in predicting and monitoring Ebola outbreaks has gained attention in recent years. Various studies have developed models to assist in early detection, prediction, and response to EVD outbreaks.

One study by \citet{siettos2015modeling} focused on modeling the 2014 Ebola virus epidemic in West Africa using agent-based simulations and temporal analysis. The researchers characterized and predicted epidemic dynamics in Liberia and Sierra Leone, and emphasize the impact of interventions and human behavior on controlling the virus spread. Similarly, researchers in Beijing in China utilized large-scale machine learning in an artificial society to predict the Ebola outbreak by reconstructing an artificial Beijing based on geodemographics and optimizing behaviors using machine learning techniques. This study involved building Ebola propagation models based on West African parameters to analyze the epidemic scenarios and propose interventions \cite{zhang2015large}.

In another significant effort, machine-learning prognostic models were developed using patient data from the 2014-16 outbreak in Liberia and Sierra Leone. These models, which incorporated variables such as viral load and age, were designed to predict the likelihood of death. An application called Ebola Care Guidelines was also created to provide access to evidence-based supportive care guidelines and highlight key prognostic signs and symptoms \cite{colubri2019machine}. 

ProPublica developed a machine learning model to predict future Ebola outbreaks using satellite image data on forest loss. The model identified at-risk areas in Africa based on factors such as wildlife presence, forest fragmentation, and population density, and the model revealed regions resembling previously affected places \cite{Hwang2023machine}. Concurrently, \citet{pigott2014mapping} mapped the zoonotic niche of EVD in Africa using species distribution models. This study compiled data on zoonotic transmissions, infections in bats and primates, and environmental covariates, and predicted a zoonotic transmission niche across 22 countries in Central and West Africa.

In a systematic comparison of different modeling approaches, \citet{viboud2018rapidd} highlighted the performance of various models in predicting synthetic Ebola outbreak scenarios. Ensemble predictions based on a Bayesian average of multiple models outperformed individual models and null auto-regressive models. This finding emphasized the utility of simple reactive models with few parameters for short-term incidence predictions. Complementing these efforts, \citet{buscema2020analysis} used three distinct algorithms—Topological Weighted Centroid (TWC), Dynamic Naive Bayesian (DNB) Algorithm, and Selfie Artificial Neural Network (ANN)—to analyze Ebola outbreaks. These methods accurately predicted the origins and diffusion of outbreaks, confirmed by WHO data.

\citet{wong2017systematic} reviewed early modeling studies of the 2013-2016 Ebola epidemic in West Africa, focusing on Sierra Leone, Liberia, Guinea, and Nigeria. The review included 41 studies using various modeling approaches to estimate epidemiological parameters such as the basic reproductive number ($R_0$), serial interval, latent period, infectious period, and case fatality rate. Median $R_0$ values ranged from 1.30 to 1.84 in Sierra Leone, Liberia, and Guinea, with a notably higher $R_0$ of 9.01 in Nigeria.

\citet{chretien2015mathematical} provided an extensive review of 66 mathematical modeling studies published during the 2013-2015 West Africa Ebola epidemic. The study addressed uncertainties such as the basic reproduction number ($R_0$), intervention effectiveness, epidemic forecasting, regional spread, phylogenetics, and vaccine trial feasibility. In a similar vein, \citet{tshomba2022development} aimed to improve decision-making for isolating suspected EVD cases at triage points during outbreaks. Using data from the Democratic Republic of the Congo (DRC) EVD outbreak, they developed and evaluated Clinical Prediction Scores (CPS and ECPS), with the ECPS showing superior performance (ROC AUC of 0.88) compared to CPS (ROC AUC of 0.71).

Forecasting efforts included \citet{carias2019forecasting}, who analyzed 26 manuscripts published between 2014 and April 2015, focusing on models forecasting the Ebola outbreak in Liberia, Guinea, and Sierra Leone. Forecasted case counts varied widely, with more accurate predictions for short-term periods. Additionally, \citet{wang2015ebola} provided real-time estimations of critical parameters for the Ebola outbreak in West Africa using a numerical fitting model. They estimated $R_0$ values and final outbreak sizes for Guinea, Liberia, and Sierra Leone.

\citet{hranac2019predicting} investigated the correlation between bat birthing cycles and EVD spillover events, using ensemble niche models to predict spatio-temporal variations in African bat birthing patterns. This study identified significant correlations between bat birth cycles and EVD spillover risk. \citet{loubet2016development} aimed to create a predictive score for EVD using clinical and epidemiological factors, demonstrating an AUC of 0.85 for predicting laboratory-confirmed EVD.

\subsubsection{Measles surveillance}\label{measles}

Machine learning has demonstrated promising results in predicting measles cases in Africa. For instance, in Ghana, a study by \citet{gyebi2023prediction} compared the performance of five machine learning techniques—Naive Bayes, Random Forest, Decision Trees, Support Vector Machines, and Artificial Neural Networks—against traditional statistical methods for predicting measles cases. The Random Forest model exhibited superior performance with the highest sensitivity (0.88), specificity (0.96), and overall accuracy (0.92). This highlights the potential of machine learning in enhancing measles surveillance and prediction.

In Tanzania, a descriptive study of measles surveillance data from 2018 to 2022 emphasized the critical need for robust surveillance and routine immunization to achieve the required immunity levels to interrupt measles outbreaks \citep{michael2024trends}. The study did not employ AI techniques but underscored the potential for applying machine learning to similar datasets to improve predictive capabilities.

The World Health Organization (WHO) African Region has implemented case-based surveillance guidelines for measles and rubella. These guidelines provide a framework for data collection and analysis, which could be significantly improved by incorporating AI-driven predictive models.\footnote{\textit{WHO African Regional measles and rubella surveillance guidelines.} WHO African Region. Retrieved from \url{https://www.afro.who.int/publications/who-african-regional-measles-and-rubella-surveillance-guidelines-0}. Accessed on July 1, 2024.}

Further, a study in Uganda examined the effectiveness of the measles surveillance system and its positive predictive value (PPV). The analysis revealed that out of 6,974 suspected measles cases, 943 (14\%) were confirmed through IgM serology. However, only 72\% of Integrated Disease Surveillance and Response (IDSR) alerts had blood samples collected for laboratory confirmation, below the WHO-recommended standard of 80\% \citep{nsubuga2017positive}. The PPV of the case-based surveillance system was 8.6\%, which indicates that many suspected cases were not true measles cases.

A dynamic model combining clinical and diagnostic data was developed to simulate measles surveillance and estimate the reporting rate and annual incidence of measles under varying vaccination coverages. This model demonstrated that integrating clinical surveillance with diagnostic confirmation, even at limited testing rates (20\%), resulted in unbiased estimates of measles incidence and reporting rates for moderate vaccination levels \citep{leung2024combining}. For instance, with a 40\% vaccination coverage, the estimated reporting rate was 7.7\% with 20\% testing, compared to a true rate of 10\%.

In Somalia, vaccination campaigns conducted since 2018 were assessed using a transmission model based on comprehensive epidemiological data, including case-based surveillance, vaccine registries, and serological surveys. The study estimated that, without the 2019–2020 campaigns, there would have been over 10,000 additional deaths \citep{thakkar2024estimating}; this underscores the impact of vaccination campaigns in reducing measles susceptibility and mortality.

Social media data have also been leveraged for measles surveillance. An analysis of 157,334,195 tweets revealed that 36,816,202 tweets (23.4\%) originated from Africa, highlighting the continent's significant role in the context of measles outbreaks. Countries like Egypt, Sudan, Congo, South Africa, Guinea, and Morocco were major sources of tweets related to measles \citep{jahanbin2024using}. This data aligns with WHO reports indicating a substantial rise in measles cases worldwide, particularly in regions with inadequate healthcare systems.

An Artificial Neural Network (ANN) approach was employed to analyze and forecast the child immunization rate against measles in Djibouti from 1982 to 2030. The study used an ANN (12, 12, 1) model with a hyperbolic tangent activation function and found the model to be stable, indicated by low error metrics (Error: 0.006882, MSE: 0.108126, MAE: 0.208483). Projections suggested that child immunization against measles in Djibouti would remain around 90\% per year over the next decade \citep{maradze2021modelling}.

In the Democratic Republic of Congo, a study assessed the impact of measles immunization activities on measles incidence from 2010 to 2013. Using data from the Integrated Disease Surveillance and Response system, the study found that the presence of supplementary immunization activities (SIAs) and outbreak response immunization (ORI) campaigns in the previous year was associated with a significant decrease in measles incidence \citep{doshi2015effect}.

In Southwest Nigeria, an analysis of 10,187 suspected measles cases revealed seasonal peaks in infections, with an increasing trend projected until 2015, assuming sustained intervention efforts. The study emphasized the need for continuous vaccination efforts to control measles \citep{fatiregun2014epidemiology}. In Niger, a study using a stochastic metapopulation model found that measles outbreaks were highly episodic due to powerful seasonality in transmission rates. Increased vaccine coverage, although insufficient for local elimination, resulted in highly variable major outbreaks \citep{ferrari2008dynamics}.

In Guinea, real-time modeling of a measles outbreak in Lola following the Ebola crisis in 2015 applied multiple statistical methods to estimate the susceptible population and projected future measles incidences. Despite initial forecasts predicting a large outbreak, subsequent vaccination campaigns mitigated the outbreak, highlighting the importance of timely interventions \citep{graham2018challenges}.

A state-space model for measles transmission incorporating particle filtering and maximum likelihood estimation was presented, demonstrating unbiased estimates of unobserved disease burden and underreporting rates. The model accurately predicted future measles cases in simulations based on data from Niger, Ethiopia, and the Democratic Republic of the Congo \citep{eilertson2019estimation}.

In Ethiopia, the performance of a Genetic Algorithm (GA) based trained Recurrent Fuzzy Neural Network (RFNN) was compared to an Adaptive Neuro-Fuzzy Inference System (ANFIS) in predicting monthly measles cases. The GA-based RFNN outperformed ANFIS, with lower error metrics across training, testing, and overall datasets \citep{uyar2019forecasting}.

In Nigeria, a deterministic mathematical model was developed to study measles transmission dynamics, and highlighted how increasing both vaccination and hospitalization rates can reduce the overall burden of measles \citep{james2022mathematical}. In Kano, Nigeria, a study examined the relationship between weather variables and measles incidence, finding significant linear effects of maximum and minimum temperatures, as well as relative humidity, on measles occurrence. The best-fitting model for predicting measles incidence was a combination of Poisson regression with all weather variables and ARIMA \citep{akinbobola2018predicting}.

An analysis of measles epidemiology in 40 African countries revealed a significant correlation between vaccination coverage and the age distribution of measles cases, emphasizing the need for high vaccination coverage to control and eventually eliminate measles \citep{goodson2011changing}.

A review of measles epidemiology in low-income countries gathered data essential for creating dynamic models of measles transmission. The review noted significant variations in age-stratified case reporting and seroprevalence studies \citep{szusz2010review}. In Uganda, a dynamic stochastic model integrated with an economic cost model evaluated the cost-effectiveness of SIAs from 2010 to 2050. The analysis found that SIAs were a highly cost-effective public health intervention \citep{bishai2011cost}.

In Cameroon, the spatial dynamics of the 2012 measles outbreak were characterized using multivariate time-series models. These models revealed significant spatial heterogeneity in vaccination coverage and transmission patterns \citep{parpia2020spatio}. 

In Kenya, a geospatial information system platform combined with a measles compartment model evaluated the health and economic benefits of a two-dose measles immunization outreach strategy for hard-to-reach children. The study found this strategy to be cost-effective at varying outreach coverages \citep{lee2019economic}.

\subsubsection{Tuberculosis detection}\label{tuberculosis}

We analyze a wide array of studies focused on the application of machine learning (ML) and deep learning (DL) techniques for tuberculosis (TB) detection across various contexts, predominantly within Africa. This analysis draws on multiple research efforts, highlighting their methodologies, key findings, and the potential implications for public health in high TB burden regions.

In Uganda, \citet{sekandi2023application} used deep learning frameworks to evaluate self-recorded medication intake videos from TB patients. The 3D ResNet model showed the best performance, with a sensitivity of 94.57\%, specificity of 54.57\%, F1-score of 92.3, precision of 90.2\%, and an AUC of 0.87. This study highlights the efficiency of deep learning models in real-time monitoring of TB medication adherence.

\citet{qin2024computer} evaluated 12 computer-aided detection (CAD) products using data from a South African TB prevalence survey. The best-performing products, Lunit and Nexus, achieved AUCs near 0.9, and demonstrated diverse accuracy among CAD products in detecting bacteriologically confirmed TB. The study emphasizes the importance of selecting tailored thresholds for different population characteristics to optimize CAD product usage.

A review by \citet{oloko2022systematic} of 62 articles concluded that CAD systems utilizing DL, particularly convolutional neural networks (CNNs), significantly improve TB diagnosis accuracy. The use of public datasets like Montgomery and Shenzhen ensures broad validation across different demographics, and underscores the potential of these systems in high-burden regions.

\citet{kuang2022accurate} developed ML models to predict drug resistance in \textit{Mycobacterium tuberculosis} (MTB) using data from 10,575 MTB isolates. The 1D CNN model outperformed traditional methods, achieving F1-scores ranging from 81.1\% to 98.2\% for various anti-TB drugs. This study highlights the role of ML in providing accurate and stable predictions for TB drug resistance. Similar approaches could be beneficial for addressing drug resistance in Africa.

In Mukono district, Uganda, \citet{gichuhi2023machine} explored individual risk factors for TB treatment non-adherence using five classification algorithms. The SVM model achieved the highest accuracy at 91.28\%, and identified key predictive factors such as TB type, GeneXpert results, and antiretroviral status. This study demonstrates the potential of ML techniques in identifying patients at risk of non-adherence and suggests their use in targeted interventions.

A deterministic mathematical epidemic model developed by \citet{oshinubi2023mathematical} to study the impact of TB in East Africa highlights the importance of vaccination and treatment strategies. The model's simulations showed that increasing vaccination and treatment availability significantly reduces TB prevalence, and supported the effectiveness of public health measures in controlling TB outbreaks.

\citet{siamba2023application} applied ARIMA and hybrid ARIMA models to predict TB incidences among children in Kenya. The hybrid ARIMA-ANN model demonstrated superior predictive accuracy and indicated significant under-reporting of TB cases among children. This study underscores the potential of hybrid models in improving TB surveillance and response strategies.

\citet{ojugo2021multi} developed a decision-making framework for TB diagnosis and prediction using Bayesian networks. The study, based on data from Nigeria, demonstrated high predictive capability with an accuracy rate of 93.76\%. This framework highlights the potential of Bayesian models in processing large datasets to predict TB cases effectively.

In Uganda, \citet{shete2017evaluation} assessed the utility of measuring antibody responses to 28 \textit{M. tuberculosis} antigens for TB screening. The study found that a panel of eight antigens achieved a sensitivity of 90.6\% and specificity of 88.6\%, and suggested that serological assays based on these antigens could be a useful tool for TB screening in high-burden settings. Integrating AI techniques into this type of screening could potentially further improve the accuracy of the tests by analyzing larger datasets and identifying subtle patterns. 

The predictive performance of logistic regression and regularized ML methods for TB in people living with HIV was compared by \citet{oduor2023comparison} in Kenya. Elastic net regression outperformed traditional logistic regression, demonstrating better predictive performance and highlighting the potential of regularized ML methods in TB prediction.

A study conducted by \citet{peng2024explainable} at the Chongqing Public Health Medical Center in China assessed treatment outcomes for patients with both TB and diabetes mellitus using ML techniques applied to electronic medical records. This study, which included 429 patients, identified nine key predictors using the random-forest-based Boruta algorithm, with drug resistance being the most significant. The XGBoost model outperformed other models with an area under the curve (AUC) of 0.9281, indicating high predictive accuracy. This approach could be replicated in Africa to improve TB treatment outcomes by leveraging electronic medical records and machine learning.

Similarly, \citet{hansun2023machine} conducted a systematic literature review encompassing 47 studies on TB detection using ML and DL methods. The datasets most frequently used included the Shenzhen and Montgomery County datasets. DL architectures like ResNet-50, VGG-16, and AlexNet, along with ML methods such as SVM and random forests, demonstrated high accuracy. The pooled sensitivity and specificity of these methods were 0.9857 and 0.9805, respectively, demonstrating a strong potential for accurate TB detection. These findings suggest that similar methodologies could be applied in Africa to enhance TB detection.

\citet{balakrishnan2023machine} reviewed the use of ML approaches in diagnosing TB via biomarkers. The review included 19 studies from countries such as Tanzania, the Philippines, and Mozambique, focusing on supervised learning methods. SVM and RF algorithms emerged as the top performers, with SVM achieving up to 97.0\% accuracy, 99.2\% sensitivity, and 98.0\% specificity. The studies predominantly utilized protein-based biomarkers, with some exploring gene-based biomarkers, highlighting the promise of ML in TB diagnosis in low-resource settings.

In an effort to optimize feature extraction and selection from TB-related images, \citet{hrizi2022tuberculosis} proposed a model using a genetic algorithm (GA) and an SVM classifier. Using the ImageCLEF 2020 dataset, the model achieved up to 97\% accuracy for specific TB classification tasks. This underscores the model's robustness in identifying TB manifestations, demonstrating the importance of feature selection and parameter optimization in enhancing classification performance. This model could be adapted for use in Africa to improve TB diagnosis accuracy. 

\citet{balzer2020machine} developed risk scores for identifying high-risk individuals for HIV acquisition in rural Kenya and Uganda using ML approaches. The machine learning risk score demonstrated superior efficiency and sensitivity, and highlighted the potential of ML algorithms like Super Learner in optimizing HIV prevention strategies in generalized epidemic settings.

\citet{moradinazar2023epidemiological} assessed the TB burden across 21 countries in the Middle East and North Africa (MENA) region using the Global Burden of Disease 2019 database. The study found significant reductions in TB metrics over 29 years, with Afghanistan having the highest incidence and DALYs rates. This study underscores the need for targeted health policies and interventions to address TB in high-burden countries. These methodologies could also be replicated in sub-Saharan Africa to reduce TB incidence and mortality.

In Algeria, \citet{asma2021analysis} used a hierarchical multicriteria analysis method integrated with a GIS database to identify and map the vulnerability of various communes to TB. The study found a significant correlation between TB cases and socio-economic conditions, highlighting the disease's resurgence due to deteriorating living conditions.

\citet{kilale2022economic} assessed the financial impact of TB on patients in Tanzania, finding that nearly half of the TB-affected households faced catastrophic costs. The study emphasizes the need for collaborative efforts across health, employment, and social welfare sectors to minimize costs and improve treatment adherence. AI could help identify risk factors and develop more targeted interventions to reduce costs and improve patient outcomes.

In Cape Town, South Africa, \citet{andrews2013modeling} estimated the risk of TB transmission on public transit using a modified Wells-Riley model. The study found that minibus taxis had the highest risk of TB transmission due to poor ventilation and crowding, and highlighted the public health challenge posed by poorly ventilated public transit environments.

\citet{houben2016feasibility} explored the feasibility of achieving WHO TB reduction targets in high-burden countries. The study found that a combination of interventions tailored to country-specific needs shows promise, particularly in South Africa, where significant reductions in TB incidence and mortality could be achieved with combined interventions.

In South Africa, \citet{pretorius2014potential} investigated the impact of expanding antiretroviral therapy (ART) eligibility and coverage on TB incidence and mortality. The study projected that expanding ART eligibility to all HIV-positive individuals could significantly reduce TB incidence and mortality, highlighting the importance of collaborative HIV-TB programs.

A comprehensive modeling approach by \citet{ojo2023integer} to predict TB incidence in African countries (using data from 2000 to 2021) found that TB incidence rates varied significantly across the continent. The study highlights the diverse epidemiological landscapes in Africa and the importance of tailored public health interventions to control TB.

In the Democratic Republic of Congo, \citet{van2008counseling} evaluated three models of provider-initiated HIV counseling and testing for TB patients. The study found that HIV testing at TB clinics or primary health care centers led to higher uptake compared to referral to freestanding VCT clinics, and emphasized the need for integrated HIV-TB care.

A mathematical model developed by \citet{kasereka2020analysis} to study TB transmission dynamics in the DRC highlighted the role of rapidly evolving latent cases in increasing TB incidence. The study suggests that reducing contact and transmission rates can achieve TB eradication with effective public health measures.

\citet{hane2007identifying} reported low TB cure rates and high treatment default rates in Senegal and identified several problems at different levels of healthcare. The study emphasizes the importance of addressing access to diagnosis and treatment facilities, communication, and patient follow-up to improve TB control.

An economic evaluation by \citet{d2023cost} assessed the impact of decentralizing TB diagnosis services for children to district hospitals and primary health centers in Cambodia, Cameroon, Côte d'Ivoire, Mozambique, Sierra Leone, and Uganda. The study found that decentralizing to district hospitals could be cost-effective in some countries, while decentralizing to primary health centers generally was not. For example, the incremental cost-effectiveness ratios (ICERs) for decentralizing to district hospitals ranged from \$263 in Cambodia to \$342 in Côte d'Ivoire per disability-adjusted life year (DALY) averted. This study provides valuable insights for optimizing resource allocation to improve TB diagnosis and treatment among children.

\citet{seri2017prevalence} estimated the prevalence of pulmonary TB among inmates in Côte d'Ivoire and found a significantly higher TB prevalence than in the general population. The study highlights the need for systematic annual TB screening campaigns in prisons to address the high disease burden. In Abidjan, Côte d'Ivoire, \citet{bouscaillou2016prevalence} assessed the health needs of people who use drugs (PWUD) and found high HIV and TB prevalence among specific subpopulations. The study underscores the urgent need for targeted interventions to improve HIV and TB prevention and healthcare access for PWUD.

\citet{iradukunda2021key} identified key factors associated with multidrug-resistant TB (MDR-TB) in Burundi, including residence in rural areas, previous TB treatment, and close contact with MDR-TB patients. The study's model demonstrated high predictive accuracy, highlighting the importance of addressing these risk factors to control MDR-TB.

In Tanzania and Ethiopia, \citet{jonathan2024machine} developed ML models to predict the performance of trained TB-detection rats. The SVM model achieved the highest accuracy of 83.39\% and demonstrated the potential of incorporating additional diagnostic variables to enhance predictive performance.

The utility of qXR AI software\footnote{qXR software, developed by Qure.ai in Mumbai, India, is an AI-based tool used to analyze chest radiographs by generating threshold scores that range from 0 to 1 for particular abnormalities.} in detecting lung cancer and pulmonary TB in South Africa was assessed by \citet{nxumalo2024utility}. The software showed high sensitivity and specificity for both conditions and highlighted its potential to enhance diagnostic accuracy in resource-constrained healthcare environments.

\citet{glaser2023incidental} examined the detection of non-TB abnormalities using CAD systems in high TB-burden regions. The study found that CAD software could identify a range of non-TB conditions, emphasizing the need for enhanced AI software to ensure comprehensive patient care.

A review by \citet{scott2024diagnostic} investigated the utility of CAD software in TB screening in a community-based setting in Africa and found a pooled sensitivity of 0.87 and specificity of 0.74. The review highlights the potential of CAD as a cost-effective TB screening tool in resource-limited settings but notes concerns regarding bias and applicability.

\citet{david2024tweaking} explored the complexities of implementing CAD algorithms for TB in different global contexts. The study highlights the need for local calibration and adjustments to improve the accuracy and utility of CAD tools, particularly in resource-limited environments.

\citet{ngosa2023assessment} identified and estimated the prevalence of non-TB abnormalities in digital chest X-rays (CXRs) with high CAD4TB scores among individuals without bacteriologically confirmed TB. This cross-sectional analysis included participants aged 15 years and older from 21 communities in Zambia and South Africa, who had high CAD4TB scores ($\leq$70) but no TB confirmation, no history of TB, and not on TB treatment. The study found that 46.7\% of the reviewed CXRs displayed non-TB abnormalities, with pleural effusion/thickening/calcification and cardiomegaly being the most common.

\begin{tcolorbox}[colback=purple!10!white, colframe=blue!30!gray, title=Top Performing Models]

Figure \ref{fig:model_performance_histo_disease} shows an overall picture of the best performing models in disease detection and prediction.  For HIV detection, \citet{mutai2021use} developed a XGBoost model and achieved an F1 score of 0.92. \citet{ahmad2024can} reported a remarkable accuracy of 0.9962 for Cholera Outbreak Risk Prediction (CORP). \citet{colubri2019machine} developed prognostic models for Ebola, achieving a ROC AUC of 0.85. \citet{gyebi2023prediction} utilized a Random Forest model for measles detection, reaching an accuracy of 0.92, while \citet{gichuhi2023machine} employed a Support Vector Machine (SVM) model for tuberculosis detection, with an accuracy of 0.91.
\end{tcolorbox}

\begin{table*}[h!]
\centering
\fontsize{5}{5}\selectfont
\renewcommand{\arraystretch}{1.2}
\caption{Summary of AI Applications in Real-time Surveillance and Reporting}
\begin{tabular}{|C{2cm}|C{2.5cm}|C{1cm}|C{2cm}|C{2.cm}|C{3.5cm}|}
\hline
\textbf{Authors} & \textbf{Model} & \textbf{Disease} & \textbf{Country} & \textbf{Application} & \textbf{Case Study Description} \\ \hline

\citet{cheng2020applying} & Random Forest, XGBoost & Influenza & Taiwan & ILI Trend Prediction & Accurate prediction of ILI trends using machine learning, showcasing effectiveness of these methods. \\ \hline
\citet{nsoesie2021forecasting} & RF, SVM & Influenza & Cameroon & ILI Trend Forecasting & Evaluation of Google search queries to forecast ILI trends, achieving high predictive performance with SVM. \\ \hline
\citet{jang2021effective} & LSTM & Influenza & Global & Influenza Trend Prediction & Use of LSTM models to predict influenza trends using news data. \\ \hline
\citet{yang2023deep} & Multi-attention LSTM & Influenza & Megacity & Influenza Trend Prediction & Developed MAL model integrating heterogeneous data to predict influenza trends. \\ \hline
\citet{olukanmi2021utilizing} & LSTM, RF, ARIMA & Influenza & South Africa & ILI Trend Forecasting & Forecasting ILI trends using Google search data combined with machine learning and time series modeling. \\ \hline
\citet{steffen2012improving} & - & Influenza & Sub-Saharan Africa & Surveillance Improvement & SISA project aimed at improving influenza surveillance by establishing sentinel sites and data collection mechanisms. \\ \hline
\citet{jiang2018mapping} & BPNN, GBM, RF & Zika Virus & Central Africa & Epidemic Mapping & Mapping probability of Zika epidemic outbreaks globally, identifying high-risk regions. \\ \hline
\citet{akhtar2019dynamic} & Dynamic Neural Network & Zika Virus & Americas & Geographic Spread Prediction & Real-time prediction of Zika outbreak spread incorporating epidemiological and socioeconomic data. \\ \hline
\citet{messina2016mapping} & Boosted Regression Trees & Zika Virus & Global & Environmental Suitability Mapping & Created high-resolution global map indicating suitability for Zika virus transmission. \\ \hline
\citet{olaniyi2018dynamics} & SIR Model & Zika Virus & Global & Transmission Modeling & Analysis of Zika virus transmission dynamics incorporating mosquito-related parameters. \\ \hline
\citet{okyere2020analysis} & Nonlinear Control Problem & Zika Virus & Global & Optimal Control Strategies & Studied effectiveness of combined controls (e.g., personal protection, vaccination) in reducing Zika virus spread. \\ \hline
\citet{caldwell4612733vector} & Genetic Variation Model & Zika Virus & Africa & Outbreak Risk Projection & Investigated mosquito genetic variation and climate factors influencing Zika virus transmission patterns. \\ \hline
\citet{barhoumi2022overcoming} & Nowcasting Framework & COVID-19 & Sub-Saharan Africa & Economic Activity Tracking & Developed framework to predict real-time economic activity using machine learning during the pandemic. \\ \hline
\citet{chimbunde2023machine} & ANN, RF & COVID-19 & South Africa & ICU Mortality Prediction & Identified predictors of COVID-19 ICU mortality using machine learning models. \\ \hline
\citet{mansell2023predicting} & ML Techniques & COVID-19 & Africa & SAHOs Prediction & Predicting issuance of COVID-19 stay-at-home orders across 54 African countries using machine learning. \\ \hline
\citet{abegaz2022artificial} & ANFIS, FFNN, SVM, MLR & COVID-19 & East Africa & Mortality Prediction & AI-driven ensemble model to predict COVID-19 mortality, comparing several ML models. \\ \hline
\citet{onovo2020using} & Lasso Regression, EBK & COVID-19 & Sub-Saharan Africa & Outbreak Risk Analysis & Identified key factors associated with COVID-19 outbreaks using spatial analysis. \\ \hline
\citet{ileperuma2023predicting} & ML Models & Malaria & Senegal & Prevalence Prediction & Predicted malaria prevalence based on rainfall patterns using ML models. \\ \hline
\citet{mariki2022combining} & Random Forest & Malaria & Tanzania & Malaria Diagnosis & High accuracy in diagnosing malaria using demographic data and clinical symptoms. \\ \hline
\citet{hemachandran2023performance} & CNN, MobileNetV2, ResNet50 & Malaria & Global & Blood Smear Analysis & Compared deep learning models for detecting malaria in blood smears, achieving high accuracy. \\ \hline
\citet{nkiruka2021prediction} & XGBoost & Malaria & Sub-Saharan Africa & Climate-based Prediction & Classified malaria incidence based on climate variability, achieving high accuracy. \\ \hline
\citet{masinde2020africa} & Gradient Boosted Trees & Malaria & Africa & Outbreak Prediction & Evaluated ML algorithms for malaria prediction using historical data. \\ \hline
\citet{shuaib2018avadar} & AVADAR System & Poliovirus & Nigeria & AFP Detection & Increased detection and reporting of AFP cases using machine learning and smartphone video analysis. \\ \hline
\citet{kamadjeu2009tracking} & GIS, Spatial Analysis & Poliovirus & Various & Case Mapping & Mapped polio cases, vaccination coverage, and population movements using AI and GIS. \\ \hline
\citet{dougherty2019paper} & Digital Elevation Modeling & Poliovirus & Various & Environmental Surveillance & Improved poliovirus detection in sewage using machine learning and environmental data. \\ \hline
\citet{khan2020novel} & K-means Clustering & Poliovirus & Pakistan & Outbreak Prediction & Analyzed spatio-temporal spread of polio using sales and travel data to predict outbreaks. \\ \hline
\citet{schaible2019twitter} & LDA Modeling & Poliovirus & Various & Media Analysis & Analyzed polio-related tweets and media articles to identify thematic differences and public engagement. \\ \hline
\citet{njoroge2023use} & AI/ML Models & Mental Health & Kenya & Mental Health Prediction & Deployed mobile app platform to identify predictors of mental health disorders among healthcare workers. \\ \hline
\citet{alharahsheh2021predicting} & Voting-Ensemble, SVM, RF & Mental Health & Kenya & Depression Prediction & Used various ML algorithms to predict depression, achieving high performance with the Voting-Ensemble model. \\ \hline
\citet{ugar2024designing} & - & Mental Health & Sub-Saharan Africa & Diagnosis Complexity & Addressed complexities and cultural considerations in using AI/ML for diagnosing mental health disorders. \\ \hline
\citet{kutcher2019creating} & - & Mental Health & Malawi, Tanzania & Youth Mental Health & Developed a horizontally integrated model to improve mental health care and literacy among youth in low-resource settings. \\ \hline
\citet{kemp2020patient} & - & Mental Health & South Africa & Depression Detection & Investigated integration of depression treatment into primary care, highlighting predictors of detection and referral. \\ \hline

\end{tabular}
\label{tab:real_time_surveillance}
\end{table*}

\subsection{Real-time surveillance and reporting} \label{realtime}
AI-powered tools are revolutionizing real-time health surveillance in Africa by continuously monitoring data and generating alerts when unusual patterns are detected. This advancement is crucial in a region where delays in data reporting can significantly hinder timely responses to health threats. For influenza surveillance, machine learning and deep learning approaches have shown great promise (\S\ref{influenza}). For instance, machine learning methods like random forests and support vector machines have been effectively used to forecast influenza-like illness (ILI) trends in African countries such as Cameroon and South Africa \cite{nsoesie2021forecasting}. Studies demonstrate that integrating Google search data with historical ILI data can improve the accuracy of predictions, highlighting the potential of machine learning techniques to improve early warning systems in Africa \cite{olukanmi2021utilizing}. Additionally, projects like the Strengthening Influenza Sentinel Surveillance in Africa (SISA) have established sentinel sites and improved data collection and reporting mechanisms, significantly boosting the region's surveillance capacity \cite{steffen2012improving}.

In the realm of vector-borne diseases, AI has also been instrumental in monitoring and predicting the spread of Zika virus and malaria (\S\ref{zika}, \S\ref{malaria}). Machine learning models, such as backward propagation neural networks (BPNN) and random forests, have been used to map Zika epidemic outbreaks, identifying high-risk areas in Central Africa \cite{jiang2018mapping}. Similarly, in malaria reporting, ML models have been developed to predict malaria prevalence based on climatic factors, achieving high accuracy in various regions across Africa \cite{ileperuma2023predicting,masinde2020africa}. These models facilitate timely interventions by forecasting outbreaks and identifying high-risk zones. For poliovirus surveillance, AI-driven tools like the Auto-Visual AFP Detection and Reporting (AVADAR) system have significantly improved the detection and reporting of acute flaccid paralysis (AFP) cases \cite{shuaib2018avadar} (\S\ref{polio}). Geographic information systems (GIS) combined with AI algorithms further improve real-time mapping and analysis of polio cases, vaccination coverage, and population movements, optimizing resource allocation for vaccination campaigns \cite{kamadjeu2009tracking,dougherty2019paper}.

In the case of COVID-19, machine learning techniques have been applied to predict mortality rates, the issuance of stay-at-home orders, and the impact of public health interventions, demonstrating significant improvements in pandemic response \cite{chimbunde2023machine, abegaz2022artificial} (\S\ref{covid}). For mental health monitoring (\S\ref{mental}), AI-driven tools have been used to identify predictors of mental health disorders, enabling rapid diagnostics and precise interventions, particularly among healthcare workers and vulnerable populations \cite{njoroge2023use, alharahsheh2021predicting}. 

\subsubsection{Influenza surveillance} \label{influenza}

Influenza surveillance in Africa has gained increased attention in recent years, with researchers exploring artificial intelligence and other computational approaches to improve traditional surveillance systems. Several studies have demonstrated the potential of these methods for forecasting influenza-like illness (ILI) trends and improving early warning capabilities.

\begin{figure}[t!]
    \centering
    \includegraphics[width=0.45\textwidth]{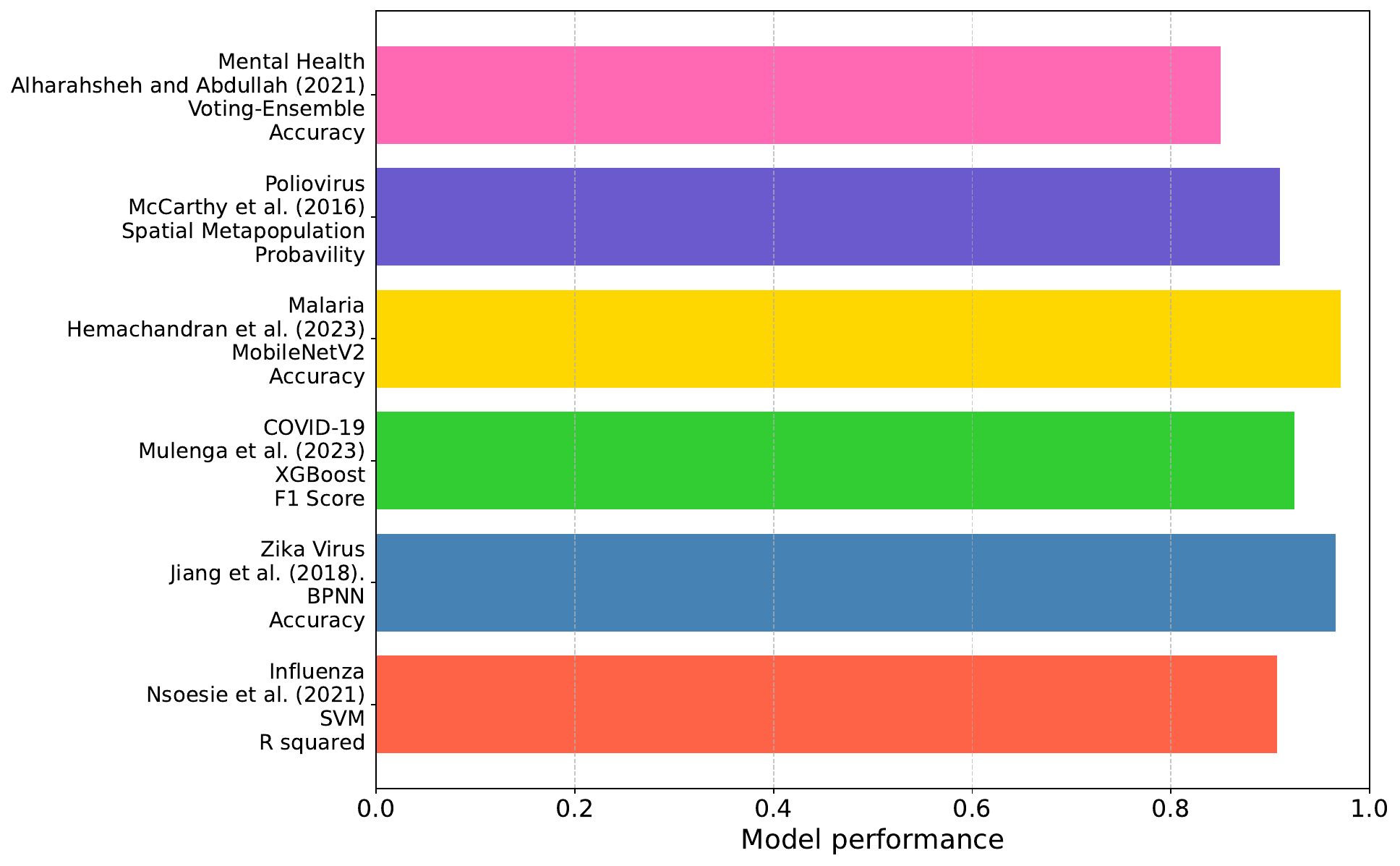} 
    \caption{Performance metrics of best models per disease for Real-time surveillance and reporting. Note that {\it y-axis} indicates {\it Disease + Authors + Model + Metric}.}
    \label{model_performance_histo_real_time_surv}
\end{figure}

\citet{cheng2020applying} utilized various machine learning methods, including random forest and extreme gradient boosting, to accurately predict ILI trends in Taiwan. Although their work is not specific to Africa, it showcases the effectiveness of machine learning for influenza forecasting. In the African context, \citet{nsoesie2021forecasting} evaluated the utility of Google search queries to forecast trends of ILI in Cameroon from 2012 to 2018 and applied multiple statistical and machine learning models, including RF and SVM regression. The RF and SVM models showed the highest predictive performance, with SVM achieving an R\textsuperscript{2} of 0.877 and RF an R\textsuperscript{2} of 0.781. The study highlighted significant correlations between specific search terms and ILI cases, with out-of-sample predictions demonstrating that SVM had the lowest mean RMSE.

The analysis revealed regional trends, with the East and Adamaoua regions having the highest number of reported cases. At the national level, SVM regression had a mean RMSE of 1.078 and an R\textsuperscript{2} of 0.877, while RF regression had a mean RMSE of 1.41 and an R\textsuperscript{2} of 0.781. One-week-ahead forecasts indicated a correlation of 72.9\% for SVM and 66.8\% for RF between predicted ILI values and the reported rate of ILI per 100,000 individuals. This demonstrates the applicability of machine learning techniques in African settings.

Deep learning approaches have also shown promise in influenza surveillance. \citet{jang2021effective} demonstrated that long short-term memory (LSTM) models could accurately predict influenza trends using news data. Building on this, \citet{yang2023deep} developed a multi-attention LSTM deep learning model (MAL model) that integrated heterogeneous data from multiple sources to predict influenza trends in a megacity, capturing different influenza incidence signals within the population. Similarly, \citet{olukanmi2021utilizing} utilized Google search data in combination with deep learning, machine learning, and time series modeling to forecast ILI trends in South Africa. Their comparison of various methods, including LSTM networks, random forests, and ARIMA models, showed that models integrating Google Trends and historical ILI data outperformed those using either data source alone.

Despite these advancements, challenges remain in applying these methods across Africa, primarily due to the lack of comprehensive surveillance data in many countries. The SISA (Strengthening Influenza Sentinel Surveillance in Africa) project aimed to improve influenza surveillance in sub-Saharan Africa by establishing sentinel sites and improving data collection and reporting mechanisms \cite{steffen2012improving}. This project involved countries like Angola, Cameroon, Ghana, and Nigeria, leading to the development of national surveillance protocols and training programs, significantly improving the region's surveillance capacity.

Contextualized approaches are crucial when using digital data for disease surveillance in sub-Saharan Africa. \citet{nsoesie2021forecasting} emphasized considering cultural and societal practices unique to the region when selecting search terms for models. Their study demonstrated that Google Trends data alone can serve as a reliable proxy for ILI surveillance in South Africa, with the feedforward neural network (GT-FNN) model achieving the lowest root mean squared error (RMSE) of 13.44 and mean absolute error (MAE) of 9.87 for nowcasting. Support vector machine (GT-SVM) also performed well with a Pearson correlation coefficient of 0.9068.

The SISA project, detailed by \citet{steffen2012improving}, improved national influenza surveillance capabilities in eight African countries, including Angola, Cameroon, Ghana, Nigeria, Rwanda, Senegal, Sierra Leone, and Zambia. The project led to the initiation of epidemiological influenza data collection, with Sierra Leone identifying 1235 ILI cases and 282 severe acute respiratory infections (SARI) cases between August and December 2011, of which 12.7\% tested positive for influenza virus RNA.

\citet{nsoesie2014systematic} reviewed various methodologies for forecasting influenza outbreaks, finding that while diverse approaches are capable of predicting outbreak measures, variability in accuracy and data quality presents challenges. The key results included correlation coefficients for forecast accuracy ranging from 58\% to 93.5\% and RMSE values between 0.47 and 0.61.

\citet{budgell2015evaluation} compared the Viral Watch (VW) and SARI surveillance systems in South Africa from 2009 to 2012. The VW system had a higher influenza detection rate (mean 41\%) compared to the SARI system (mean 8\%), with better timeliness in specimen collection but lower data quality.

\citet{mejia2019leveraging} explored the use of Google search data combined with traditional epidemiological data to predict influenza activity in Africa. The ARGO model, adapted for South Africa, Algeria, Morocco, and Ghana, outperformed traditional autoregressive models and Google Trends alone, demonstrating improved prediction accuracy with lower RMSE values. In South Africa, ARGO showed a 59\% lower RMSE compared to Google Flu Trends (GFT).

\citet{bokonda2021machine} investigated machine learning methods for predicting various outbreaks, including influenza, with SVM achieving the highest accuracy at 90.42\%. \citet{shah2024seasonal} presented a machine learning model based on AdaBoost to predict the antigenic properties of influenza A (H3N2) virus isolates. The model achieved an MAE of 0.702 and strong discriminatory ability in identifying antigenic variants.

\citet{katz2012influenza} and \citet{caini2018epidemiology} provided extensive analyses of influenza seasonality and epidemiology in Africa, and they highlighted the increased surveillance efforts and the significant contributions of international organizations. \citet{n2016effects} demonstrated the importance of climatological parameters, such as rainfall, in predicting influenza trends in Abidjan and suggested their integration into predictive models for effective public health planning.

\citet{nabakooza2022molecular} investigated the genomic evolution of influenza A viruses (IAVs) in Africa and underlined the need for expanded surveillance and routine whole-genome sequencing. Finally, the WHO's efforts to establish and improve influenza diagnostic capacities across African countries were detailed by \citet{world2017influenza}, emphasizing the importance of continuous and comprehensive surveillance systems to better understand and mitigate the impact of influenza in the region.

\subsubsection{Zika virus monitoring}\label{zika}

Zika virus (ZIKV) is a mosquito-borne virus that has caused global outbreaks in recent years, primarily in Central Africa, Southeast Asia, the Pacific Islands, and Central and South America \cite{alves2018research}. The spread of Zika virus poses a significant public health challenge, and monitoring its transmission is crucial for effective control and prevention strategies. Recent studies have employed machine learning techniques to improve Zika virus monitoring and surveillance efforts \cite{akhtar2019dynamic}.

Accurate prediction of Zika virus outbreaks is critical for optimal resource allocation in vector control efforts. Statistical modeling and machine learning techniques have been utilized to predict the occurrence and arrival time of arboviral infections, including dengue fever, which shares similarities with Zika virus in terms of transmission dynamics \cite{akhtar2019dynamic}. These models help guide vector control strategies by identifying high-risk areas and informing the deployment of preventive measures.

To increase data availability on the epidemiology of Zika virus circulation in Africa, seroepidemiological studies have been conducted in West Africa. These studies evaluate the immunity to Zika virus in selected cohorts over specific time periods and provide valuable information on Zika virus exposure and transmission in the region \cite{marchizika}. For instance, \citet{jiang2018mapping} employed multiple machine learning models, including backward propagation neural networks (BPNN), gradient boosting machines (GBM), and random forests (RF), to map the probability of Zika epidemic outbreaks globally. Their results identified Central Africa as one of the high-risk regions for Zika transmission, with the BPNN model achieving the highest predictive accuracy with a 10-fold cross-validation area under the curve (AUC) of 0.966, followed closely by GBM and RF models. This research demonstrated the potential of machine learning approaches in identifying vulnerable areas in Africa.

Moreover, \citet{akhtar2019dynamic} developed a dynamic neural network model to predict the geographic spread of Zika outbreaks in real-time, incorporating epidemiological data, passenger air travel volumes, vector habitat suitability, and socioeconomic factors. Although their study focused on the Americas, the methodology could be adapted for African contexts, maintaining an average accuracy above 85\% for prediction windows up to 12 weeks. Similarly, \citet{alexander2022using} used machine learning to understand microgeographic determinants of \textit{Aedes aegypti}, the primary vector of Zika virus, in Miami-Dade County. Their methodology of using random forest models to analyze neighborhood-level differences in mosquito populations could be applied to African urban areas, emphasizing the importance of local context in vector control strategies.

Further, \citet{messina2016mapping} created the first high-resolution global map of environmental suitability for Zika virus transmission to humans using an ensemble of 300 boosted regression trees. The results indicated that over 2.17 billion people globally live in areas suitable for Zika virus transmission, with the highest risk regions in tropical and subtropical zones, including Brazil, Nigeria, and India. The model achieved high predictive performance with an AUC of 0.829 and underlined the significant factors influencing Zika virus transmission such as annual cumulative precipitation and temperature suitability for dengue virus transmission.

In the context of intervention strategies, \citet{olaniyi2018dynamics} formulated and analyzed a Zika virus transmission model incorporating three nonlinear forces of infection. Their results emphasized the sensitivity of the basic reproduction number ($R_0$) to mosquito-related parameters, suggesting that mosquito control is crucial for managing Zika virus spread. Optimal control strategies, including the use of insecticide-treated bed nets, condoms, routine checks, treatment, and mosquito-reduction measures, were found to be effective in reducing the number of infected individuals and mosquitoes.

\citet{okyere2020analysis} developed a nonlinear optimal control problem to study Zika virus transmission, incorporating five time-dependent control functions: personal protection, condom use, vaccination, treatment, and insecticide spraying. Numerical simulations using the Runge-Kutta method demonstrated that the combination of these controls significantly reduces the spread of Zika virus, with the combination of personal protection, condom use, and treatment showing a substantial decline in Zika cases. The cost-effectiveness analysis revealed that the strategy combining vaccination, treatment, and insecticide spraying is the most cost-effective.

\citet{obore2019zika} reviewed the epidemiological evidence and distribution of Zika virus in Africa, highlighting its origins, spread, determinants, complications, and management strategies. They identified critical determinants of Zika virus spread, including climate, sociodemographic factors, and human density, and emphasized the importance of improving surveillance mechanisms and vector control. Over 87 countries reported Zika virus presence by 2019, with significant cases in Cabo Verde and Angola.

Additionally, \citet{caldwell4612733vector} investigated factors influencing Zika virus transmission across Africa, focusing on mosquito genetic variation and climate. Their model, validated with historical patterns, projected future outbreak risks in 59 urban centers and underlined the significant role of mosquito genetics in transmission patterns. The study predicted that about one-third of Africa's largest cities are currently suitable for Zika virus transmission, with another third becoming suitable by the end of the century.

\citet{mpeshe2017modeling} presented a simple SIR (Susceptible-Infected-Recovered) model to analyze Zika virus transmission dynamics in Africa, computing the basic reproduction number ($R_0$) and performing sensitivity analysis. The sensitivity analysis indicated that $R_0$ is most sensitive to the natural death rate of \textit{Aedes} mosquitoes, and suggested that increasing the death rate of mosquitoes and improving human recovery can effectively control the disease.

To improve real-time surveillance, \citet{teng2017dynamic} developed a dynamic forecasting model for Zika virus using real-time online search data from Google Trends. Their ARIMA model, incorporating Google Trends data, demonstrated strong predictive capabilities, closely matching actual reported data during the early November 2016 Zika virus epidemic.

\citet{majumder2018seasonality} explored the potential link between seasonal patterns of birth defects and congenital Zika syndrome in a West African hospital. They found a marked increase in birth defects between March and July, correlating with high mosquito activity periods. This study, conducted in a medium-risk West African country, underscores the importance of understanding seasonal dynamics in Zika virus transmission.

\citet{quadri2017targetzika} presented a framework for disaster situation modeling in smart cities through micro-reporting from mobile phones. This framework, although applied in Brazil, offers a robust tool for disease surveillance and control that could be replicated in African regions facing similar epidemics. \citet{romero2023modelling} highlighted the significance of understanding the epidemiological dynamics of vector-borne diseases through mathematical and statistical models. The review acknowledged Zika virus's long period of circulation in Central Africa and emphasized its continued prevalence.

Furthermore, \citet{marchi2020zika} evaluated the prevalence of Zika virus antibodies in subjects from Mali, Senegal, and The Gambia. Their findings indicated active Zika virus circulation in Senegal and The Gambia, with a strong association between Zika virus prevalence and increasing age. \citet{shen2016phylogenetic} investigated the genetic evolution and migration pathways of Zika virus through phylogenetic analysis. They identified significant roles of Senegal and Côte d'Ivoire in Zika virus evolution. These insights are crucial for understanding the geographic origins and global spread of Zika virus, informing prevention and control measures.

\subsubsection{COVID-19 tracking}\label{covid}

The COVID-19 pandemic has posed significant challenges globally, particularly in Africa, where data scarcity and limited healthcare resources have exacerbated the crisis. Artificial intelligence has emerged as a powerful tool to address these challenges by providing real-time tracking, prediction, and decision-making support. Studies have demonstrated the application of machine learning (ML) across various domains to improve pandemic response and management in African countries.

One notable study by the International Monetary Fund developed a machine learning framework to track real-time economic activity in sub-Saharan Africa. The framework, known as ``nowcasting,'' uses indicators available earlier than official GDP statistics to predict economic activity months or quarters in advance. This approach bridges the gap between policymaking and data availability, providing timely information crucial for effective policy responses during the pandemic. The study shows that machine learning techniques can extract valuable signals from sparse data, significantly improving the timeliness and accuracy of economic assessments in the region \cite{barhoumi2022overcoming}.

In the healthcare domain, predicting COVID-19 mortality in intensive care units (ICUs) has been a critical application of ML in Africa. \citet{chimbunde2023machine} revealed several critical insights into the determinants of COVID-19 ICU mortality in South Africa. The semi-parametric logistic regression model identified significant predictors, including advanced age, severe symptoms, low oxygen saturation, and the presence of asthma. The study highlighted that asthmatic patients had six times higher odds of mortality. Performance metrics for the machine learning models were promising, with the artificial neural network model achieving an accuracy of 71\%, precision of 83\%, F1 score and MCC of 100\%, and recall of 88\%, demonstrating robust predictive capability. In comparison, the random forest model provided a recall of 76\%, precision of 87\%, and MCC of 65\%. Both models consistently identified age, intubation status, cluster, diabetes, and hypertension as top predictors of mortality.

Machine learning has also been applied to predict the issuance of COVID-19 stay-at-home orders (SAHOs) in Africa. Researchers applied machine learning techniques to predict the issuance of SAHOs in 54 African countries \cite{mansell2023predicting}. The results revealed significant predictors, including temporal diffusion, previous adoption by other countries, median age, cumulative COVID-19 cases per doctor, land area, death rate from air pollution, urban population percentage, time required to start a business, and the percentage of the population adhering to Islam. Temporal diffusion emerged as the strongest predictor, indicating that countries were influenced by the actions of their peers. The model's accuracy improved to 78\% with these variables, and it demonstrated the value of machine learning in uncovering complex relationships in health policy-making.

Additionally, \citet{abegaz2022artificial} proposed an AI-driven ensemble model to predict COVID-19 mortality in East Africa. Their study compared several models, including adaptive neuro-fuzzy inference system (ANFIS), feedforward neural network (FFNN), support vector machine (SVM), and multiple linear regression (MLR), finding that the ANFIS ensemble model performed best with a coefficient of determination of 0.9273, effectively boosting the predictive performance of individual AI-driven models.

Machine learning applications extend to virus detection, spread prevention, and medical assistance. \citet{onovo2020using} employed Lasso regression and Empirical Bayesian Kriging (EBK) to identify and spatially analyze key factors associated with COVID-19 outbreaks in sub-Saharan Africa. Utilizing data from Johns Hopkins University and additional socio-demographic and health indicators, the study identified seven significant predictors of COVID-19 risk, including new HIV infections, pneumococcal conjugate-based vaccine, incidence of malaria, and diarrhea treatment. The geospatial analysis demonstrated increasing COVID-19 spread daily with significant clustering in South Africa and potential increases in West African countries.

\citet{ibrahim2023multi} evaluated the performance of various ML models across multiple African countries. The study found that in North Africa, particularly in Morocco, the SVM model achieved the highest validation performance, while ANFIS outperformed other models in Sudan. In East Africa, ANFIS demonstrated superior accuracy in Rwanda, and in West Africa, Nigeria's predictions were more accurate using AI models, especially ANFIS. In Southern Africa, ANFIS performed well in Namibia and South Africa, with the highest validation accuracy observed in Central Africa's Cameroon using ANFIS.

AI and big data have been leveraged to improve COVID-19 public health responses and vaccine distribution in Africa. The Africa-Canada Artificial Intelligence and Data Innovation Consortium (ACADIC) developed a deep neural network (DNN) model to prioritize vaccination for vulnerable populations in Gauteng Province, South Africa. The model demonstrated that vaccinating 20\% of the adult population could reduce severe COVID-19 illness by over 80\% \cite{mellado2021leveraging}.

\citet{whata2021machine} developed a hybrid model combining long-short term memory auto-encoder (LSTMAE) and kernel quantile estimator (KQE) to detect change-points in population mobility data, quantifying the causal effects of lockdown measures on various locations. The study found significant decreases in mobility across all categories, with the most notable reductions in grocery and pharmacy, workplaces, retail and recreation, transit stations, parks, and residential areas, all with p-values less than 0.05.

Further, \citet{busari2022modelling} provided a detailed comparison of different forecasting models for predicting new COVID-19 cases in Nigeria. The inverse regression model outperformed other regression models, with ARIMA (4, 1, 4) standing out among ARIMA models. For machine learning models, the Fine tree algorithm emerged as the best performer, achieving an R\textsuperscript{2} of 0.90 and an RMSE of 0.22165, demonstrating superior forecasting accuracy.

Innovative uses of technology have been observed across different African nations in response to the pandemic. Ethiopia and Sierra Leone implemented internet-based payment systems and electronic passes for physical distancing, while drones were used in South Africa, Morocco, Sierra Leone, and Tunisia for public health messaging and disinfection \cite{maharana2021covid}. Countries like Senegal and Uganda promoted hygiene through locally designed contactless soap dispensers and solar-powered handwashing sinks. Mass communication technologies were employed, with Kenya's Ushahidi platform gathering information and offering help during lockdowns, and Guinea using SMS messaging and caller ringtones for information dissemination.

In the realm of disease surveillance, dashboards were developed to monitor COVID-19 cases, tests, and deaths. Tunisia used Tableau dashboards, Rwanda implemented a GIS system, and South Africa repurposed technology for tracking COVID-19 clusters. Mobile applications were developed for recording personal information and temperature in Ethiopia, and contact tracing applications were used in Ghana and Tunisia. Genomic epidemiology played a significant role, with Ghanaian, Zambian, and Gambian scientists utilizing genomic sequencing for tracking virus mutations and infection origins. Innovative approaches such as sample pooling and home-grown swab tests were employed in disease diagnosis to increase testing capacity and reduce costs.

To improve pandemic preparedness, \citet{mulenga2023predicting} employed various ML models to predict mortality in hospitalized COVID-19 patients in Zambia. The XGBoost model achieved the best performance with an accuracy of 92.3\%, recall of 94.2\%, F1-score of 92.4\%, and ROC AUC of 97.5\%, identifying key features such as length of stay, white blood cell count, age, and underlying health conditions as significant predictors of mortality.

Predictive models were also used to estimate confirmed cases and deaths due to COVID-19 in Africa. \citet{likassa2021predictive} employed several models, including cubic and quadratic algorithms, revealing that the cubic model outperformed others with an R\textsuperscript{2} of 0.996 and an F-value of 538.334, showing significant regional variations in COVID-19 spread and impact.

\citet{mathaha2022leveraging} explored the impact of COVID-19 on Botswana and South Africa, using AI to identify high-risk population groups and assist in prioritizing vaccine distribution. The AI model, a deep neural network, classified patients based on health status, and emphasized the elderly who accounted for 70\% of major COVID-19 comorbidities like hypertension and diabetes.

\citet{maurice2021covid} aimed to predict the trajectory of COVID-19 infections in Kenya using the ARIMA model. The study identified ARIMA(0,1,2) as the best fit, projecting a constant increase in cases before the curve eventually flattens, which aids in resource planning and management.

Using data from Cheikh Zaid Hospital, \citet{laatifimachine} combined biological and non-biological data to identify key indicators of COVID-19 severity. They achieved 100\% accuracy using classifiers like XGBoost, AdaBoost, Random Forest, and ExtraTrees. \citet{madani2021using} evaluated various ML algorithms for detecting fake COVID-19 news on Twitter and found that the Random Forest algorithm achieved the highest accuracy of 79\%.

\citet{dodoo2024using} identified barriers to COVID-19 vaccine uptake in Ghana, using machine learning algorithms to highlight significant predictors such as age, facility type, marital status, and health insurance status. The random forest model achieved the highest predictive accuracy with an AUC of 0.82, demonstrating its effectiveness in identifying factors influencing vaccine uptake.

\citet{diallo2022artificial} developed clinical predictive models in Senegal to estimate the likelihood of COVID-19 infection using basic patient information, achieving an accuracy of 73\% and an AUROC of 69\%. The best-performing models were XGBoost and ANN, demonstrating the utility of AI in clinical prediction.

In Senegal, \citet{ndiaye2020visualization} applied various ML techniques for forecasting COVID-19 and found that the MLP and Prophet models provided the most accurate forecasts. In Egypt, \citet{marzouk2021deep} used LSTM and CNN models to predict COVID-19 cases, with LSTM achieving superior accuracy. Similarly, \citet{amar2020prediction} analyzed COVID-19 data in Egypt using regression models, with the fourth-degree polynomial model providing the best fit for predicting future cases.

Beyond genomic epidemiology, ML was also utilized to address the infodemic during the COVID-19 pandemic. For instance, Mutanga et al. (2022) analyzed 68,000 tweets related to COVID-19 in South Africa, revealing significant findings such as the prevalence of conspiracy theories linking 5G technology to the virus spread and public reactions to alcohol sales bans. The study underscored the importance of sentiment analysis in detecting fake news and highlighted the role of social media in shaping public perceptions during the pandemic.

\subsubsection{Malaria reporting}\label{malaria}

In recent years, the application of machine learning (ML) techniques in malaria reporting and prediction has gained significant traction, demonstrating their potential in enhancing disease surveillance, diagnosis, and control strategies across various regions in Africa. Researchers in Senegal developed ML models to predict malaria prevalence based on rainfall patterns, utilizing monthly and province-wise datasets of recorded malaria cases from 2009 to 2021, along with satellite-derived rainfall estimates \cite{ileperuma2023predicting}. This approach aimed to forecast the number of malaria cases for the next three months across different regions and emphasized the importance of selecting suitable ML algorithms for effective prediction modeling.

In Tanzania, a study focused on diagnosing malaria using data from the Morogoro and Kilimanjaro regions. A random forest ML model achieved high accuracy rates of 95\% in Kilimanjaro, 87\% in Morogoro, and 82\% across the entire dataset, based on demographic data and clinical symptoms \cite{mariki2022combining}. This highlights the regional specificity in malaria prediction, which can inform targeted intervention strategies.

The performance of three deep learning models—CNN, MobileNetV2, and ResNet50—was compared using a dataset of 27,558 blood smear images from the NIH. The MobileNetV2 model achieved the highest accuracy at 97.06\%, followed by ResNet50 and CNN. ResNet50 also demonstrated superior precision (0.97), recall (0.97), and ROC AUC value (96.73), which indicated a robust ability to distinguish between infected and uninfected cells \cite{hemachandran2023performance}.

Researchers have also leveraged climate data for malaria prediction. A model developed by \citet{nkiruka2021prediction} classified malaria incidence based on climate variability across six Sub-Saharan African countries using the XGBoost algorithm. The model achieved high accuracy and AUC scores, with mean AUC scores of 0.97 for Mali, 0.94 for Cameroon, 0.91 for DRC, 0.97 for Nigeria, 0.94 for Niger, and 0.92 for Burkina Faso. The integration of feature engineering and k-means clustering significantly improved the model's accuracy and underscored the role of climate variables in malaria transmission.

In another comprehensive study, \citet{masinde2020africa} evaluated nine ML algorithms using historical climate and malaria incidence data in Africa, with Gradient Boosted Trees achieving the highest overall rank. The developed malaria predictor system, leveraging deep learning, achieved a prediction accuracy of up to 99\%. These findings underscore the potential of ML techniques in providing accurate malaria outbreak predictions and enhancing early warning systems.

In Burkina Faso, a data-driven early warning system for predicting malaria epidemics demonstrated notable accuracy and reliability. The developed algorithm maintained a precision of 30\% and a recall rate of over 99\% for low epidemic alert thresholds, and precision improved significantly to over 99\% for high alert thresholds \cite{harvey2021predicting}. Incorporating rainfall data further improved prediction accuracy.

Machine learning has also been applied to classify different clinical malaria outcomes using haematological data. In Ghana, six different ML approaches were evaluated, with an artificial neural network emerging as the best model. The multi-classification model achieved over 85\% accuracy, and binary classifiers identified key haematological parameters distinguishing uncomplicated malaria from non-malarial infections and severe malaria from non-malarial infections with high accuracy \cite{morang2020machine}.

Sea surface temperature (SST) variability has been investigated as a predictive tool for malaria outbreaks. Researchers found significant correlations between SST variability and malaria outbreaks, particularly in the Indian Ocean and western Pacific Ocean. ML models trained with these climatic indices and past malaria cases achieved prediction accuracies of up to 80\% for a lead time of up to 9 months \cite{martineau2022predicting}.

In Zimbabwe, various ML models, including logistic regression and random forest, were evaluated for predicting malaria incidences based on environmental factors. Logistic regression and random forest both achieved the highest prediction accuracy at 83\% and demonstrated the efficacy of these models in guiding resource allocation and early warning system development \cite{mbunge2022application}.

An analysis of environmental factors influencing the biting rates of malaria vectors in Burkina Faso revealed significant correlations between biting rates and meteorological and landscape variables. Random forest models showed good predictive accuracy, with PR-AUC values of 0.56 for \textit{An. funestus}, 0.46 for \textit{An. gambiae s.s.}, and 0.60 for \textit{An. coluzzii} \cite{taconet2021data}.

In Northern Benin, an intelligent malaria outbreak warning model using climatic variables was developed. The SVM algorithm performed the best, predicting 82\% of malaria incidence. The model projected an increase in malaria incidence from 2021 to 2050 under different climate scenarios \cite{gbaguidi2024towards}.

The effectiveness of various machine learning algorithms for identifying malaria parasites in blood smear images was investigated, with the Fine Gaussian SVM exhibiting the highest true positive rate (99.8\%) for detecting plasmodium, though its overall accuracy was lower at 82.0\%. Subspace k-nearest neighbors achieved the best overall performance with an accuracy of 86.3\% \cite{olugboja2017malaria}. 

In Nigeria, the Region-specific Elastic-Net based Malaria Prediction System (REMPS) was developed, achieving MAE $\leq 0.06$ and MSE $\leq 0.007$, with over 80\% of predictions falling within a clinically relevant error tolerance range \cite{brown2020data}. This system demonstrates the potential of supervised ML techniques in predicting malaria prevalence with high accuracy.

The use of deep learning CNN approaches to predict the geographic origin of malaria infections using whole genome sequence data from \textit{Plasmodium falciparum} and \textit{Plasmodium vivax} showed over 90\% accuracy at the country level and highlighted the potential of ML in supporting malaria control and elimination efforts \cite{deelder2022geographical}.

\subsubsection{Poliovirus surveillance}\label{polio}

Poliovirus surveillance has been significantly improved by the integration of artificial intelligence across Africa, leading to improved detection and reporting systems. One notable innovation is the Auto-Visual AFP Detection and Reporting (AVADAR) system, which employs machine learning algorithms to analyze video footage captured by community health workers using smartphones. Trained to recognize signs of acute flaccid paralysis (AFP), such as limb weakness or paralysis in children under 15, AVADAR generates automatic alerts for rapid investigation and response by health authorities \cite{shuaib2018avadar}. The implementation of AVADAR in Nigeria, for instance, resulted in a substantial increase in AFP case detection and reporting \cite{shuaib2018avadar}.

Geographic information systems (GIS) and spatial analysis are also utilized to map and visualize polio cases, vaccination coverage, and population movements in real-time. These technologies, integrated with AI algorithms, enable researchers to identify high-risk areas, predict potential outbreaks, and optimize resource allocation for vaccination campaigns \cite{kamadjeu2009tracking, dougherty2019paper, oteri2021application, gammino2014using, thompson2020review}. For example, environmental surveillance has benefited from AI through the use of Digital Elevation Modeling and machine learning techniques to improve poliovirus detection in sewage and wastewater samples. This improves early warning capabilities and allows for targeted interventions in areas where the virus may be circulating undetected \cite{hamisu2022characterizing}.

AI-driven initiatives have demonstrated promising results. In Nigeria, AVADAR significantly increased the detection of AFP cases, and the use of GIS and AI-powered spatial analysis improved the targeting of vaccination campaigns and the identification of under-immunized populations \cite{shuaib2018avadar, dougherty2019paper}. Additionally, the WHO Regional Office for Africa supports the development and implementation of AI in polio eradication efforts through small research grants for innovative projects leveraging AI and machine learning for polio surveillance and public health applications.\footnote{ Retrieved from \url{https://polioeradication.org} and \url{https://www.afro.who.int/health-topics/polio}. Accessed on July 7, 2024.}

In Nigeria, the study by \citet{hamisu2022characterizing} aimed to improve environmental surveillance (ES) sensitivity for poliovirus detection by evaluating various site characteristics. Data collected over four quarterly visits to 78 ES sites across 21 states revealed significant variability in enterovirus detection rates, ranging from 9\% to 100\%. High-performing sites were associated with larger catchment populations, higher total dissolved solids, and elevated pH levels. The correlation between catchment population estimates derived from digital elevation models (DEM) and GRID3 data with enterovirus detection highlighted the reliability of publicly available population data. Predictive models using random forests demonstrated that water quality data, field team observations, and ES officer estimates could predict site performance with a median accuracy of 75\%.

Machine learning models have also been employed to predict polio outbreaks. \citet{khan2020novel} developed a model integrating data from the National Institute of Health (NIH), medical store sales records, and transport logs to analyze the spatio-temporal spread of polio. The K-means clustering algorithm identified potential outbreaks based on increased sales of polio-related medicines and travel patterns of affected individuals, showing the model's effectiveness in early detection of high-risk regions.

The study by \citet{schaible2019twitter} analyzed polio-related tweets and media articles across India, Iraq, Nigeria, Pakistan, and Syria using latent Dirichlet allocation modeling. The analysis revealed thematic differences between Twitter and traditional media. Tweets from India predominantly celebrated its polio-free status, while those from Pakistan focused on eradication efforts and challenges due to political instability. Traditional media, however, emphasized negative themes like violence and conflicts and suggested that social media might be more favorable for promoting health campaigns and engaging global audiences positively.

The S2M model by \citet{zhou2016s2m} examined the genetic state transitions of poliovirus from the Sabin-1 vaccine strain to the Mahoney wild type, incorporating parameters like replication error rate and recombination rate. Simulations demonstrated the complex interplay of these factors in driving genetic transitions and highlighted the model's utility in understanding poliovirus evolution.

In Nigeria, the use of satellite data has significantly improved polio vaccination campaigns. \citet{borowitz2023examining} detailed how accurate GIS maps created from satellite data identified over 3,000 previously unmapped settlements, increasing vaccination coverage in Kano state from 60\% in 2011 to 89\% in 2014. This approach resulted in substantial economic benefits, estimated at \$46.0 million to \$153.9 million in one year, primarily from reduced vaccination campaign efforts and health benefits of preventing polio cases.

In South Africa, \citet{waggie2012randomized} conducted a trial comparing monovalent type 1 (mOPV1) and type 3 (mOPV3) oral poliovirus vaccines with trivalent oral poliovirus vaccine (tOPV) in newborns. The results showed higher seroconversion rates for mOPV1 and mOPV3 compared to tOPV, supporting their use in supplemental immunization activities in regions where these poliovirus types circulate.

A statistical model developed by \citet{o2011statistical} analyzed wild poliovirus outbreaks in Africa from 2003 to 2010, identifying significant predictors such as migration, immunization coverage, and child mortality rates. The model demonstrated 82\% accuracy in forecasting outbreaks six months in advance, with better immunization coverage associated with shorter and smaller outbreaks.

\citet{kalkowska2021global} updated a global differential equation-based poliovirus transmission model to incorporate recent data and programmatic experiences. Despite progress, the model indicated ongoing wild poliovirus (WPV1) transmission and the emergence of vaccine-derived poliovirus (cVDPV2), and emphasized the need for improved immunization coverage and improved outbreak response strategies.

In conflict-affected regions like Borno and Yobe, \citet{kalkowska2021modeling} used a deterministic differential equation-based model to assess poliovirus transmission dynamics. The findings indicated that recent immunization efforts have significantly reduced the risk of undetected transmission, although continuous improvement is needed to minimize re-emergence risks.

The spatial metapopulation model by \citet{mccarthy2016spatial} analyzed WPV1 transmission in Kano State, Nigeria, showing a 91\% probability of elimination by October 2015. This model, along with other AI-driven initiatives, underscores the critical role of advanced technologies in enhancing polio surveillance and eradication efforts across Africa.

\subsubsection{Mental health monitoring}\label{mental}

The use of mobile technology and machine learning to identify predictors of mental health disorders has shown promising results across various African countries. In Kenya, \citet{njoroge2023use} deployed a mobile application platform that utilized artificial intelligence and machine learning to identify predictors of mental health disorders among healthcare workers. The aim was to create agile and scalable systems for rapid diagnostics and precise interventions to mitigate depression and promote a healthy healthcare workforce.

\citet{alharahsheh2021predicting} explored the use of machine learning algorithms to predict depression among individuals in Kenya using a dataset from the Busara Center, which included over 70 features related to participants' health, economic activity, financial flows, and household composition. They tested several models, including SVM, Random Forest, Ada Boosting, and Voting-Ensemble techniques, with the Voting-Ensemble model achieving the highest performance (F1-score of 0.78 and accuracy of 85\%).

The complexities of using AI and ML for diagnosing mental health disorders in sub-Saharan Africa were addressed by \citet{ugar2024designing}, who argued that cultural and value-laden judgments must be incorporated into these systems. They emphasized the need for AI and ML systems to understand local perceptions of mental health disorders, avoiding generic designs that might lead to misdiagnosis. The study highlighted challenges specific to African contexts, such as different interpretations of conditions like schizophrenia and psychosis.

Innovative, horizontally integrated models have been developed to improve mental health care for youth in low-resource settings, as demonstrated by \citet{kutcher2019creating} in Malawi and Tanzania. Their model, which aimed to improve mental health literacy (MHL) among communities, schools, and healthcare providers, utilized interactive radio programs, school-based MHL curriculum, and competency training for community healthcare providers. Results showed significant positive impacts, including increased awareness and reduced stigma around mental health issues among youth and teachers.

In KwaZulu-Natal, South Africa, \citet{kemp2020patient} investigated the integration of depression treatment into primary care and its effectiveness in detecting and referring patients for depression treatment. They found that nurses detected depressive symptoms in 50.5\% of participants, referred 36.5\% of those detected, and only 23.7\% of referred patients attended at least one counseling session. The study identified significant associations between depressive symptom severity, alcohol use severity, perceived stress, and detection.

Socioeconomic factors influencing depression were explored by \citet{folb2015socioeconomic} among 4,393 adults with chronic illnesses attending primary care clinics. Higher baseline depression scores were linked to lower education and income levels, with over 56\% of participants exhibiting significant depression symptoms. The use of antidepressant medication was more common among those with higher education and income.

The prevalence and predictors of postpartum depression (PPD) among HIV-infected and HIV-negative women were measured by \citet{mokhele2019prevalence} in a cross-sectional survey involving 1,151 women. They found that 25.0\% of women experienced PPD, with 10.4\% having major PPD. A higher proportion of HIV-negative women experienced PPD compared to HIV-positive women, with living arrangements and antenatal care attendance identified as significant predictors.

In a study of burnout and depression among surgeons, \citet{commander2020predictors} found that 48\% experienced moderate to severe depression and 38\% suffered from burnout. The key predictors included serious professional conflict, difficulty accessing childcare, and racial discrimination. African surgeons reported a more positive outlook compared to their non-African counterparts despite significant challenges.

The prevalence of suicidal ideation among South African university students was investigated by \citet{bantjes2016symptoms}, who found higher rates compared to the general population and other student populations globally. Symptoms of depression and posttraumatic stress were significant predictors of suicidal ideation.

\citet{bigna2019epidemiology} conducted a systematic review and meta-analysis on the prevalence of depressive disorders among people living with HIV (PLHIV) in Africa, incorporating data from 118 studies with 60,476 participants across 19 countries. They found that the overall prevalence of depressive disorders was 36.5\%, with probable major depressive disorders affecting 14.9\%. The study highlighted significant heterogeneity in prevalence rates, influenced by diagnostic tools, study period, and geographical distribution.

\citet{brathwaite2020predicting} explored the applicability of a depression prediction model, initially developed in Brazil, to Nigerian adolescents. The model demonstrated better-than-chance discrimination (AUC = 0.62) but was not well-calibrated, indicating the need for context-specific adjustments to improve predictive accuracy. \citet{amu2021prevalence} investigated the prevalence and factors influencing depression, anxiety, and stress among 2,456 adults in Ghana. They found that 51.8\% experienced at least one mental health issue, with significant predictors including education and income levels.

The prevalence and predictors of prestroke depression and its association with poststroke depression (PSD) in Ghana and Nigeria were examined by \citet{ojagbemi2022pre}. Prestroke depression was found in 7.1\% of participants, with significant associations identified between prestroke depression and factors such as tachycardia and low consumption of green leafy vegetables.

The effectiveness of the Beck Depression Inventory-I (BDI-I) in screening for major depressive disorder (MDD) among individuals seeking HIV testing was investigated by \citet{saal2018utility}. The BDI-I demonstrated moderate accuracy in predicting MDD and suggested the need for clinical interviews to confirm diagnoses.

The prevalence and predictors of depression among hypertensive patients in Ghana and Nigeria were assessed by \citet{ademola2019prevalence}. They found a higher prevalence of depression among Ghanaian patients (41.7\%) compared to Nigerian patients (26.6\%), with significant associations in Nigeria including younger age and concerns about medication.

AIDS-related stigmas and their impact on mental health were examined by \citet{simbayi2007internalized}, revealing significant correlations between stigmas and cognitive–affective depression among HIV-positive individuals in South Africa.

\citet{khumalo2022measuring} investigated the validity of the dual-continua model of mental health among African university students from Ghana, Kenya, Mozambique, and South Africa. Their study found support for the model and identified distinct mental health classes among students.

Depression among snakebite patients and its predictors were explored by \citet{muhammed2017predictors} in Nigeria, identifying key predictors such as snakebite complications and financial loss. The prevalence of PTSD, depression, and anxiety among Ebola Virus Disease (EVD) survivors in Beni, Democratic Republic of Congo, was assessed by \citet{kaputu2021ptsd}. Their study highlighted significant associations with factors such as gender, persistent headaches, and loss of loved ones.

\begin{tcolorbox}[colback=purple!10!white, colframe=blue!30!gray, title=Top Performing Models]
Figure \ref{model_performance_histo_real_time_surv} displays the best performances and models found in the literature for real-time health surveillance in Africa. It highlights the dominance of machine learning models over traditional methods. For example, Influenza prediction using SVM achieved an R\textsuperscript{2} of 0.9068 \cite{nsoesie2021forecasting}, and Zika Virus detection with BPNN reached a high accuracy of 0.966 \cite{jiang2018mapping}. COVID-19 detection using XGBoost reported a high F1 Score of 0.924 \cite{mulenga2023predicting}, while Malaria detection with MobileNetV2 achieved an accuracy of 0.971 \cite{hemachandran2023performance}. Poliovirus prediction using a Spatial Metapopulation model yielded a probability of 0.91 \cite{mccarthy2016spatial}, and Mental Health assessment with a Voting-Ensemble model attained an accuracy of 0.85 \cite{alharahsheh2021predicting}. These results underscore the effectiveness of machine learning models, which consistently outperform traditional methods.
\end{tcolorbox}

\section{Opportunities for AI in Africa}\label{opp_ai}

Artificial Intelligence holds immense potential to address various challenges in healthcare across Africa. AI technologies can help African nations bridge infrastructure gaps, empower local health workers, and improve community engagement. These advancements can ultimately improve health outcomes and disease management ({\bf RQ1}).

\subsection{Bridging infrastructure gaps}

One of the most significant opportunities for AI in Africa lies in addressing the infrastructure gaps that hinder effective healthcare delivery. Many regions in Africa suffer from limited access to healthcare facilities, diagnostic tools, and medical resources. AI can help bridge these gaps through innovative solutions such as telemedicine, mobile health applications, and automated diagnostic systems.

Telemedicine platforms, powered by AI, can connect patients in remote areas with healthcare providers in urban centers, enabling timely consultations and reducing the need for long-distance travel \cite{chitungo2021utility}. AI-driven diagnostic tools, such as mobile applications that analyze medical images or symptoms, can assist healthcare workers in making accurate diagnoses, even in resource-limited settings. These tools can utilize machine learning algorithms to identify diseases like malaria, tuberculosis, and HIV from images of blood smears or other diagnostic tests, ensuring early detection and treatment \cite{roche2024measuring}.

Moreover, AI can optimize the allocation of limited resources by predicting disease outbreaks and identifying high-risk areas. Predictive models that integrate data from various sources, including environmental factors, social determinants, and health records, can guide the distribution of medical supplies, vaccines, and healthcare personnel to areas where they are most needed. This proactive approach can help prevent disease spread and improve overall healthcare efficiency.

\subsection{Empowering local health workers}

AI has the potential to significantly empower local health workers by providing them with the tools and knowledge needed to deliver high-quality care. In many African countries, there is a shortage of trained healthcare professionals, and those available often lack access to continuous training and professional development opportunities. AI-driven educational platforms and decision-support systems can address these challenges \cite{karabacak2023embracing}.

AI-powered educational platforms can offer interactive training modules, simulations, and real-time feedback to healthcare workers, enhancing their skills and knowledge. These platforms can be tailored to local contexts and languages, making them accessible and relevant. Additionally, AI-driven decision-support systems can assist healthcare workers in diagnosing and managing diseases by providing evidence-based recommendations and treatment protocols \cite{denecke2024potential}.

For instance, AI algorithms can analyze patient data and suggest personalized treatment plans, considering factors such as comorbidities, genetic information, and local disease patterns. This support can improve the accuracy of diagnoses and treatment outcomes, especially in rural and underserved areas. Providing local health workers with AI tools improves the resilience and capability of healthcare systems to handle a wide range of health issues.

\subsection{Enhancing community engagement}

AI can play a crucial role in enhancing community engagement and promoting public health awareness. Effective health interventions often require community participation and behavior change, which can be challenging to achieve. AI-driven tools can facilitate communication, education, and engagement with communities, fostering a collaborative approach to health improvement.

Chatbots and virtual health assistants, powered by AI, can provide communities with reliable health information, answer questions, and dispel myths and misconceptions about diseases and treatments \cite{yang2023large}. These tools can operate in multiple languages and be accessible via mobile phones, which make them particularly useful in diverse and multilingual populations \cite{restrepo2024analyzing}.

Furthermore, AI can help design and evaluate community health programs by analyzing data on health behaviors, social networks, and cultural practices. Machine learning algorithms can identify patterns and trends in community health data, allowing public health officials to tailor interventions to specific community needs and preferences. Engaging communities in the design and implementation of health programs ensures that interventions are culturally appropriate and effective.

In addition, AI can support the monitoring and evaluation of health initiatives by providing real-time data analysis and feedback. This capability allows for the continuous improvement of programs and the identification of best practices that can be scaled up or replicated in other regions. Engaging communities through AI-driven tools can lead to more sustainable health outcomes and a greater sense of ownership and responsibility for health improvement.

\subsection{Leveraging large language models}

Large language models (LLMs) represent a significant opportunity for advancing healthcare in Africa. These models, capable of understanding and generating human language, can be instrumental in overcoming language barriers, enhancing communication, and providing tailored health information and support \cite{thirunavukarasu2023large}. They facilitate the translation of medical information and resources into various African languages, ensuring that critical health knowledge is accessible to broader populations, thereby improving health literacy and enabling individuals to make informed health decisions.

In healthcare settings, LLMs assist in documentation and record-keeping by transcribing and summarizing medical consultations, freeing up time for healthcare workers to focus on patient care. They also aid in creating personalized health education materials by generating content that addresses specific health concerns and cultural contexts \cite{thirunavukarasu2023large,clusmann2023future}. Moreover, LLMs support research and data analysis by processing large volumes of medical literature and identifying relevant studies, trends, and insights, which can accelerate the development of new treatments and interventions tailored to the unique health challenges faced by African populations \cite{karabacak2023embracing}.

The rapid advancements in LLMs have shown significant performance improvements in healthcare-specific tasks. Models like BioBERT, SCIBERT, and PubMedBERT, trained on biomedical corpora, have demonstrated superior capabilities in medical inference and disease name recognition compared to general LLMs \cite{ray2024timely}. Similarly, ClinicalBERT and GatorTron, trained on clinical notes and electronic health records, excel in understanding and processing clinical data. Conversational models like Med-PaLM and ChatDoctor, fine-tuned with medical domain knowledge, improve patient-physician interactions and diagnostic accuracy through real-time dialogue and clinical decision support systems \cite{yang2023large}.

LLMs have practical applications across various stages of the patient care journey. In preconsultation, they assist with triaging patients, recommending appropriate medical specialties, and enhancing access to care through remote diagnosis systems. During diagnosis, LLMs generate concise summaries of patient histories, support clinical decision-making, and integrate with computer-aided diagnosis (CAD) systems to interpret medical images. In patient management, they improve medication compliance and patient education by providing tailored information in simple language. Additionally, LLMs streamline administrative tasks, such as drafting discharge summaries and automating responses to patient queries, reducing the burden on healthcare professionals. Despite these promising applications, addressing challenges related to data privacy, bias, interpretability, and the role of LLMs in clinical practice is crucial to ensure safe and effective deployment \cite{yang2023large}.

The effectiveness and safety of using LLMs to assist in drafting responses to patient messages within EHR systems have been studied. For instance, at Brigham and Women's Hospital, an observational study found that LLM-assisted responses were significantly longer and provided more extensive education and self-management recommendations compared to manual responses. However, a small percentage of LLM-generated drafts posed risks of severe harm, underscoring the importance of rigorous evaluation and monitoring of LLMs in clinical settings to prevent automation bias and ensure that LLMs improve, rather than compromise, patient care \cite{chen2024effect}.

Furthermore, a Delphi study highlighted the potential impacts, both positive and negative, of implementing LLMs in healthcare. Promising use cases include supporting clinical tasks such as patient education and chronic disease management, automating documentation processes like clinical coding and summarization, and aiding in medical research and education. However, significant risks include cybersecurity breaches, patient misinformation, ethical issues, and biased decision-making. Addressing these risks through stringent data privacy regulations, quality assessment standards, and comprehensive training for healthcare professionals is essential to ensure the successful implementation of LLMs in healthcare \cite{denecke2024potential}.

\section{Challenges and Considerations}\label{challenge}

Despite the promising advancements in AI and machine learning for disease detection and prediction, several challenges and considerations must be addressed to ensure these technologies' effective and ethical implementation ({\bf RQ2}).

\subsection{Ethical, regulatory and privacy concerns}

The integration of AI in healthcare raises significant ethical, regulatory and privacy concerns. The collection, storage, and analysis of large datasets, including electronic health records, social media data \cite{xu2022unmasking}, and environmental sensors, necessitate robust data privacy measures to protect sensitive information. In many African countries, the regulatory frameworks for data protection are still developing, creating vulnerabilities that could be exploited. Ensuring patient confidentiality and securing data against breaches is paramount, especially given the potential misuse of health information \cite{alharahsheh2021predicting, ileperuma2023predicting,li2023ethics}.

Moreover, the ethical implications of AI decision-making in healthcare need careful consideration. AI algorithms can inadvertently perpetuate existing biases in healthcare data, leading to unfair treatment outcomes for marginalized populations \cite{ugar2024designing}. Transparency in AI models and the inclusion of diverse datasets are essential to mitigate bias and ensure equitable healthcare delivery. Ethical guidelines and frameworks must be established to govern the use of AI in healthcare, emphasizing accountability, fairness, and the protection of patient rights \cite{njoroge2023use}.

\subsection{Capacity building and training}

The successful deployment of AI technologies in disease detection and prediction in Africa depends significantly on local capacity building and training. Many healthcare systems in Africa face challenges such as limited access to technology, insufficient infrastructure, and a shortage of skilled healthcare professionals proficient in AI and data analytics \cite{nsoesie2021forecasting}. Addressing these gaps requires substantial investment in education and training programs to equip healthcare workers with the necessary skills to utilize AI tools effectively.

Capacity building should also focus on developing local expertise in AI and machine learning, fostering innovation, and encouraging the creation of contextually relevant solutions. Collaborations between governments, educational institutions, and international organizations can facilitate knowledge transfer and resource sharing \cite{kutcher2019creating}. Strengthening local research capabilities and promoting interdisciplinary approaches can improve the sustainability and scalability of AI-driven healthcare initiatives \cite{hamisu2022characterizing}.

\subsection{Ensuring equity and accessibility}

Ensuring equity and accessibility in the application of AI in healthcare is crucial to avoid exacerbating existing disparities. Artificial intelligence technologies must be designed and implemented to benefit all population segments, including those in remote and underserved areas \cite{ileperuma2023predicting, mariki2022combining}. This requires addressing barriers such as limited internet connectivity, inadequate healthcare infrastructure, and socio-economic inequalities that hinder access to advanced healthcare technologies \cite{masinde2020africa}.

Inclusive artificial intelligence strategies should prioritize the needs of vulnerable and marginalized groups, ensuring that interventions are culturally sensitive and tailored to local contexts \cite{ugar2024designing}. Policymakers and healthcare providers must work together to develop frameworks that guarantee equitable access to AI-driven healthcare services. This includes leveraging mobile health (mHealth) platforms and community-based approaches to reach populations with limited access to traditional healthcare facilities \cite{kamadjeu2009tracking}.

Additionally, engaging communities in the development and implementation of AI solutions can foster trust and acceptance, enhancing the effectiveness of these technologies. Public awareness campaigns and education initiatives can help demystify AI and its benefits, encouraging community participation and support for AI-driven healthcare innovations \cite{njoroge2023use}.

\subsection{Electricity limitations}
In addition to these challenges, the inconsistent and often unreliable electricity supply in many low-resource regions poses a significant barrier to the effective implementation of AI technologies \cite{streatfeild2018low,lee2022still}. Frequent power outages and lack of access to a stable electricity grid can disrupt the functionality of AI systems, hindering continuous data processing and real-time surveillance. AI models require a reliable electricity supply to be trained and to work effectively in real-time \cite{desislavov2023trends}.

\section{Discussion}\label{discussion}

The integration of artificial intelligence and machine learning (ML) in disease detection and prediction is proving to be a transformative approach in public health, especially in resource-limited settings like sub-Saharan Africa. The ability of AI to analyze large and complex datasets from various sources, including electronic health records \cite{tshimula2023redesigning}, social media, and environmental sensors, has improved the speed and accuracy of detecting disease outbreaks. This capability is crucial for timely interventions, which are vital for preventing widespread epidemics and managing existing health conditions more effectively \cite{ebulue2024developing,xu2022machine}.

One of the standout areas where AI has made significant strides is in the detection and prediction of HIV. The use of advanced algorithms such as XGBoost, Random Forest, and Artificial Neural Networks has shown high accuracy in identifying HIV-positive individuals and predicting drug resistance mutations \cite{powers2018prediction,ebulue2024machine}. These technologies not only improve the precision of diagnoses but also facilitate targeted screening interventions and personalized treatment plans. This is particularly beneficial in sub-Saharan Africa, where the burden of HIV is substantial, and healthcare resources are often limited. AI-driven models improve healthcare outcomes and support public health authorities in making informed decisions by allowing more efficient allocation of resources and better management of treatment programs \cite{tim2006predicting,roche2024measuring}.

Similarly, machine learning has been instrumental in predicting and monitoring other infectious diseases like cholera and Ebola. In the case of cholera, predictive models that integrate socioeconomic, environmental, and climatic data have achieved remarkable accuracy \cite{leo2019machine,ahmad2024can}. These models are critical for informing public health strategies, particularly in regions prone to cholera outbreaks due to inadequate sanitation and water access \cite{charnley2022drought,zheng2022cholera}. For Ebola, the use of patient data, geodemographics, and environmental factors in ML models has facilitated early detection and timely interventions, improving outbreak management \cite{siettos2015modeling,zhang2015large}. The success of these models in predicting outbreaks and assessing intervention effectiveness underscores the potential of AI in enhancing disease surveillance and response mechanisms \cite{colubri2019machine,pigott2014mapping}.

Measles surveillance has also benefited from the application of machine learning techniques. Predictive models have demonstrated the potential to improve surveillance and immunization strategies, particularly in African countries where measles remains a significant public health challenge \cite{gyebi2023prediction,michael2024trends}. Studies have shown that machine learning classifiers can outperform traditional statistical methods in predicting measles cases, leading to more robust and proactive public health interventions \cite{leung2024combining,nsubuga2017positive}. The integration of AI-driven predictive models into public health systems can significantly improve disease outbreak preparedness and response, ultimately saving lives and reducing disease burdens \cite{thakkar2024estimating,jahanbin2024using}.

AI and ML have also significantly impacted tuberculosis (TB) detection and management. Advanced models, including deep learning frameworks and computer-aided detection (CAD) systems, have shown high sensitivity and specificity in identifying TB cases. For instance, deep learning models have been used to evaluate self-recorded medication intake videos from TB patients and demonstrated high performance in real-time monitoring of medication adherence \cite{sekandi2023application}. Additionally, CAD products have been utilized in South Africa to detect bacteriologically confirmed TB with high accuracy \cite{qin2024computer}. These technologies are particularly beneficial in high TB burden regions and can enable timely diagnosis and treatment interventions, which are crucial for controlling TB outbreaks and improving patient outcomes \cite{oloko2022systematic,kuang2022accurate}.

Real-time surveillance and reporting have been revolutionized by AI-powered tools, which continuously monitor data and generate alerts when unusual patterns are detected \cite{nsoesie2021forecasting,olukanmi2021utilizing}. This advancement is crucial in Africa, where delays in data reporting can hinder timely responses to health threats. For instance, AI and ML have been effectively used to forecast influenza-like illness trends, leveraging data from Google searches and historical records \cite{yang2023deep,cheng2020applying}. The integration of such data sources has improved the accuracy of predictions, highlighting the potential of ML techniques to improve early warning systems and public health preparedness \cite{jang2021effective,mejia2019leveraging}.

Furthermore, AI has been instrumental in monitoring and predicting the spread of vector-borne diseases such as Zika virus and malaria. Machine learning models have been developed to map epidemic outbreaks, identify high-risk areas, and forecast disease prevalence based on climatic factors \cite{jiang2018mapping,akhtar2019dynamic}. These models facilitate timely interventions by predicting outbreaks and identifying high-risk zones, which is essential for effective vector control and resource allocation \cite{ileperuma2023predicting,masinde2020africa}. For poliovirus surveillance, AI-driven tools like the Auto-Visual AFP Detection and Reporting (AVADAR) system have significantly improved the detection and reporting of acute flaccid paralysis cases, demonstrating the potential of AI in enhancing surveillance and eradication efforts \cite{shuaib2018avadar,kamadjeu2009tracking}.

Overall, the integration of AI and ML in disease detection, prediction, and surveillance has shown substantial benefits in public health, particularly in African countries. More accurate and timely predictions from AI-driven models support public health authorities in efficiently allocating resources and implementing targeted interventions \cite{mayemba2024short}. The continued development and application of these technologies hold great promise for improving disease outbreak preparedness and response, ultimately contributing to better health outcomes and reduced disease burdens across the continent \cite{teng2017dynamic,bigna2019epidemiology}.

\section{Future Directions} \label{fdfuturedirection}

Looking ahead, the future of AI in public health holds immense potential for further advancements and applications. One promising avenue is the development of more sophisticated predictive models that can integrate diverse data sources, including genomics, environmental factors, and social determinants of health. These models could offer comprehensive insights into disease dynamics, enabling more precise and timely interventions. Additionally, advancements in AI interpretability, explainability and transparency will be crucial in building trust among healthcare providers and patients, ensuring that AI-driven decisions are understood and accepted by all stakeholders.



Another vital area for future exploration is the implementation of AI in personalized medicine. AI can tailor treatments to the unique genetic and environmental factors affecting each patient, improving outcomes and reducing the incidence of adverse effects. Furthermore, open-source LLM models offer significant opportunities for enhancing AI capabilities, particularly in low-resource settings such as Sub-Saharan Africa. Customizing these models to meet the specific needs of different healthcare systems can foster innovation and facilitate local implementation, leading to a more inclusive and effective global healthcare system.

\section{Conclusion}\label{conclusion}

The integration of artificial intelligence and machine learning in disease detection, prediction, and surveillance has significantly improved public health outcomes in sub-Saharan Africa by leveraging large and complex datasets from sources such as electronic health records, social media, environmental sensors, and human mobility patterns. These advancements have improved the accuracy and speed of detecting disease outbreaks, facilitating timely interventions that prevent widespread epidemics and manage existing health conditions more effectively. AI-driven models have demonstrated exceptional performance in predicting and monitoring infectious diseases like tuberculosis, HIV, cholera, Ebola, measles, Zika virus, and malaria, enabling targeted screening interventions, personalized treatment plans, and efficient resource allocation. Overall, the integration of artificial intelligence and machine learning in public health systems has shown substantial benefits, particularly in resource-limited settings, promising a future of more proactive and efficient disease management, ultimately leading to better health outcomes and reduced disease burdens across the continent.

\section*{Acknowledgments}

The authors thank all Greprovad members for helpful discussions and comments on early drafts.

\bibliography{custom}

\end{document}